 \documentclass[final,3p,times]{elsarticle}

\usepackage{graphicx}
\usepackage{float}

\usepackage{epsfig}
\usepackage{epstopdf}
\usepackage{amssymb}
 \usepackage{amsthm}
\usepackage{array}
\usepackage{lineno}
\usepackage{longtable}
\usepackage{amsmath}
\usepackage{float}
\usepackage{bm}
\usepackage{rotating}
\usepackage{tabularx}
\usepackage{mathrsfs}
\usepackage{hyperref}
\usepackage{caption}
\usepackage{picinpar}
\usepackage{wrapfig}
\usepackage{booktabs}
\usepackage{algorithm}
\usepackage{algorithmic}

\biboptions{numbers,sort&compress}

\begin{document}

\begin{frontmatter}

\title{Kernel based regression with robust loss function via iteratively reweighted least squares}

\author{Hongwei Dong}



\author{Liming Yang\corref{cor1}}
\address{College of Science, China Agricultural University, Beijing, 100083, China}
\cortext[cor1]{Corresponding author}
\ead{cauyanglm@163.com}

\begin{abstract}
Least squares kernel based methods have been widely used in regression problems due to the simple implementation and good generalization performance. Among them, least squares support vector regression (LS-SVR) and extreme learning machine (ELM) are  popular techniques. However, the noise sensitivity is a major bottleneck. To address this issue, a generalized loss function, called $\ell_s$-loss, is proposed in this paper. With the support of novel loss function, two kernel based regressors are constructed by replacing the $\ell_2$-loss in LS-SVR and ELM with the proposed $\ell_s$-loss for better noise robustness. Important properties of $\ell_s$-loss, including robustness, asymmetry and asymptotic approximation behaviors, are verified theoretically. Moreover, iteratively reweighted least squares (IRLS) is utilized to optimize and interpret the proposed methods from a weighted viewpoint. The convergence of the proposal are proved, and detailed analyses of robustness are given. Experiments on both artificial and benchmark datasets confirm the validity of the proposed methods.
\end{abstract}


\end{frontmatter}


\section{Introduction}
\label{sec:1}
\noindent Machine learning algorithms can be divided into pattern classification and regression according to the aims. For the task of pattern classification, some deep representation learning methods are quite mature and mainstream \cite{Lecun2014Backpropagation,Deep}. However, they have not shown dominance in regression problems. Least squares kernel based  regression is an effective way to this issue \cite{Audibert2011Robust,add4}, and least squares support vector regression (LS-SVR) \cite{Suykens2002Least} and extreme learning machine (ELM) \cite{Huang2006Extreme} are two representatives of this kinds of method. LS-SVR is an improved regression version of support vector machine \cite{vapniksvm}. SVM is a powerful non-parametric method for classification, whose central idea is to construct two parallel hyperplanes that separate the two classes with maximum margin. ELM is a single-hidden layer feedforward neural network, whose model parameters of hidden nodes are randomly determined. Kernel technique induced strong nonlinear capability ensures their generalization ability. And the solution obtained by solving a set of linear equations gives them good adaptability for large-scale problems, which can be regarded as a key factor for their success.
\par For the prediction of dataset $Z=\{(\bm{x}_i,y_i)\in \mathbb{R}^n\times \mathbb{R}\}_{i=1}^N$, to get a nonlinear mapping $f(\cdot)$ satisfied $f(x)\to y$, a classical paradigm of least squares kernel based methods is to pursue the structural risk minimization plus the $\ell_2$-loss empirical risk \cite{Ref16,Ref17} minimization (ERM), as follows: 
\begin{equation}
\arg\min_{f}\tau||f||^p+\frac{1}{N}\sum_{i=1}^N\ell_2(y_i-f(\bm{x}_i))\label{eq1.1}
\end{equation}
where $\ell_2(y-f(\bm{x}))=(y-f(\bm{x}))^2$. The first term is the structural risk term, which controls the model complexity to prevent overfitting. $p=2$, i.e., ridge regression, is the most widely used and effective choice. The second term is the empirical risk term, which determines the fidelity through the least square loss function. $\tau$ controls the trade-off. Both LS-SVR and ELM can be converted into the form of Eq.(\ref{eq1.1}), and the difference between them lies in the representation of $f(\cdot)$.
\par As the most widely used regressor, least squares regression focuses on minimizing the mean square error, i.e., the ERM term with $\ell_2$-loss, which is also part of the goals of LS-SVR and ELM. The reason for the effectiveness lies in that, in all Gaussian distribution induced models \cite{Catoni2010Challenging}, the empirical mean has an optimal minimax mean square error among all mean estimators. Therefore, LS-SVR and ELM also have the optimal mean of estimation for most times. Not only that, the computational convenience for large-scale problems can be guaranteed because their solutions are obtained through solving a set of linear equations. 
\par However, $\bm{x}$ is often contaminated with noise because of erroneous samplings and measurements in practical applications, and the errant observations are usually called outliers \cite{add1}. Their existence will cause the data distribution change from Gaussian to heavy-tailed, and untraceable biases will be introduced. The reason for this phenomenon can be analyzed from two aspects: From a robust statistics point of view, a bounded gradient function is needed for robust loss functions \cite{Christmann2007Consistency} and this requirement is not met by $\ell_2$-loss. From a ERM point of view, $\ell_2$-loss changes rapidly. Therefore, the loss will be very huge when the value of residual is big, so the ERM may fail to approximate the expected risk \cite{Catoni2010Challenging}. 
\par To handle this problem, robust methods have been widely studied \cite{add3}. Although they are sub-optimal estimators in theory compared with the $\ell_2$ based estimators \cite{Catoni2010Challenging,Zhang2018}, good noise robustness is their advantage. Yao and Tong proposed a generalized asymmetric $\ell_2$-loss to give different weight depending on whether the residual is positive or negative \cite{Yao1996Asymmetric}. Chen \emph{et al.} used $\ell_1$-loss for SVR and designed a effective Split-Bregman method to get the optimal solution \cite{Chen2017Least}. Some studies \cite{Mangasarian2002Robust,Chen2015A} seek robust methods based on M-estimator induced loss functions \cite{Huber1,Huber1964Robust}. Steinwart and Christmann \cite{nipssvm} made a systematic analysis about quantile and a generalized asymmetric loss was proposed to realize kernel based quantile regression. Omer \cite{Omer2017Maximum} reviewed the commonly used losses and proposed a generalized one based on maximum likelihood. Entropy has also been studied and introduced into SVR or ELM for robust regression \cite{Ren2018Correntropy,Kai2015Outlier}. Recently, to handle more general problems, truncated loss functions have been proposed and proved to be effective \cite{Yang2018,Yang2020}. The linear rising trend, even more slowly, is what these loss functions have in common, and it is also the reason why they have appealing performance. The above methods aim at modifying loss functions to acquire robustness. Besides, the improvements directly applied to ERM, i.e., truncated minimizations, have also been explored \cite{Holland2017Robust,Lugosi2016Risk,catoni2009}. Although achieved promising results, the modified objectives of truncated minimizations are really complicated. Therefore, the optimization cost of these methods is considerable. 
\par Weighted least squares is also an effective way for robust learning \cite{Suykens2002Weighted}. Generally, a weight $w\in [0,1]$ is considered for all observations. The weights of errant ones should tend to be zero so as to eliminate their adverse impacts. Although the aforementioned techniques are different in implementation, they can be regarded as obtaining robustness in an indirect way, in which the modifications of loss function and objective play the role of proxy. Many studies have proved that there is an inseparable relationship between robust loss fuctions and weighted methods \cite{Christmann2007Consistency}. On one hand, robust loss functions can be regarded as reducing the adverse effects of outliers by giving them small weights \cite{Wang2014Robust,Zhao2010Robust}. On the other hand, weighted each sample can also be considered as a special loss function \cite{chen2017neurocom}. Moreover, typical methods for solving the non-convex robust loss functions, such as difference of convex algorithm \cite{Akoa2008Combining,Yang2016A}, concave-convex procedure \cite{YUILLE2002CCCP} and half-quadratic optimization \cite{Zhang2013robust,Ran2014Half}, have been proved to be iterative variants of the weighted methods \cite{Feng2016Robust,Li2017Sparse,Xu2016Robust}. Therefore, it is possible to build a better robust loss function and solve a robust model from a weighted viewpoint \cite{Feng2016Robust}.
\par Inspired by previous works, a novel robust loss function, called $\ell_s$-loss, is proposed in this paper. On basis of $\ell_s$-loss, two alternatives to the $\ell_2$ based methods are constructed for better noise robustness. The proposed methods are optimized and interpreted by iteratively reweighted least squares (IRLS) technique \cite{Lai2013Improved,Green1984Iteratively,irlskbr}. Specifically, the objective of the proposed methods can be solved by optimizing the weighted baselines iteratively, so the suitability for large-scale problems can be guaranteed because they still solve linear systems of equations. Moreover, the effectiveness of $\ell_s$-loss can be interpreted from the analysis of the iteratively reweighted process. In addition, the convergence of the proposal are theoretically proved. Experiments on both artificial and benchmark datasets demonstrate the robustness of the proposed methods. In summary, in this paper we make the following contributions:
\begin{itemize}
	\item[-] A novel $\ell_s$ loss function is proposed to achieve better performance under the interference of noise.
	\item[-] Robust methods are constructed by combining $\ell_s$-loss and two kernel based regressors.
	\item[-] IRLS technique is used to optimize and interpret the proposed regressors.
	\item[-] The validity of proposed methods is demonstrated on artificial and benchmark datasets.
\end{itemize}
\par The rest of this paper is organized as follows: Relevant literatures are reviewed in Section \ref{sec:3}. The proposed methods are introduced in Section \ref{sec:4}. Section \ref{sec:7} discusses the robustness. In Section \ref{sec:8}, experimental results are exhibited. Conclusion and future directions are given in Section \ref{sec:6}.
\section{Background} \label{sec:3}
\noindent We start this section by introducing the least squares kernel based regression, followed by its extension, LS-SVR, ELM, and more generalized IRLS. 
\subsection{Least squares kernel based regression}
 \label{sec:3.1}
\noindent In this part, we concisely review the principles of least squares kernel based regression. Given the training set $Z=\{(\bm{x}_i,y_i)\in \mathbb{R}^n\times \mathbb{R}\}_{i=1}^N$. The commonly used $\ell_2$ based methods can be written as the following optimization problem:
\begin{equation}
\min_{f} \ \ \  \frac{1}{2} {\Vert f\Vert}_2^2+\frac{C}{2}\sum_{i=1}^N \ell_2(y_i-f(\bm{x}_i))\label{3.1.1}
\end{equation}
where $C>0$ is a regularization hyperparameter which balances the structural risk and empirical risk. For LS-SVR, the prediction function is $f(\bm{x})=\bm{w}^T\phi(\bm{x})+b$, where $\phi(\cdot):\mathbb{R}^n\rightarrow \mathbb{R}^m$ is the kernel function which maps the input space into a higher dimensional space, $\bm{w}\in \mathbb{R}^m$ is the weight vector, and $b\in \mathbb{R}$ represents the bias. The optimization problem of LS-SVR is usually converted into its dual problem by introducing Lagrangian multiplier $\bm{\alpha}$. Eliminating the original variables to obtain the solution by solving the following linear equations: 
\begin{equation}
\left[\begin{array}{cc}
    0 &    \bm{1^T}     \\
    \bm{1} &  K+\frac{1}{C}E
\end{array}\right]
\left[\begin{array}{c}
    b      \\
    \bm{\alpha}
\end{array}\right]=
\left[\begin{array}{c}
    0     \\
    \bm{Y}
\end{array}\right]\label{2}
\end{equation}
where $\bm{Y}=[y_1,y_2,\cdots,y_N]^T$, $\bm{1}=(1,1,\cdots,1)^T\in \mathbb{R}^N$, $E$ denotes the $N\times N$ identity matrix,  $K$ is the kernel matrix with $K_{ij}=\phi(\bm{x}_i)^T\phi(\bm{x}_j)$. Finally, the prediction function of LS-SVR can be written as:
\begin{equation}
f(\bm{x})=\sum_{i=1}^N\bm{\alpha}^*_iK(\bm{x},\bm{x}_i)+b^*.\label{3}
\end{equation}
\par For Tikhonov regularized ELM \cite{add2,chen2017neurocom,4938676}, $f$ is a single-hidden layer feedforward neural network. It can be written as: $f(\bm{x})=\bm{h}(\bm{x})\bm{\beta}$, where $\bm{h}(\bm{x})=(h_1(\bm{x}),h_2(\bm{x}),\cdots,h_L(\bm{x}))$, $h_i(\bm{x})$ is the hidden layer function $g(\bm{\alpha}_i,b_i,\bm{x})$ between the input layer and the $i$th hidden node ($\bm{\alpha}_i$, $b_i$ are randomly chosen). $\bm{\beta}=(\beta_1,\beta_2,\cdots,\beta_L)^T$ is the output weight between hidden and output nodes, and its solution can be expressed as:
\begin{equation}
\bm{\beta^{*}}=\left\{
\begin{array}{cc}
    H^{T}(\frac{1}{C}+HH^{T})^{-1} Y, &    N<L     \\
     (\frac{1}{C}+H^{T}H)^{-1}H^{T}Y, &  N\geq L
\end{array}\right.\label{2132}
\end{equation}
where the two forms are equivalent based on Woodbury identity \cite{Huang2006Extreme}.
\subsection{Iteratively reweighted least squares}
 \label{sec:3.3}
\noindent IRLS is a generalized robust learning paradigm, whose idea is to consider implementing multiple weighted process for better performance. For any loss function $\ell(\cdot)$, its gradient function $\psi(\cdot)$ and weight function $v(\cdot)$ are defined as:
\begin{equation}
\psi(u)=\partial\ell(u)/\partial u,\label{5}
\end{equation}
\begin{equation}
v(u)=\left\{
\begin{array}{cc}
\psi(u)/2u, &    u\neq0     \\
\psi'(0), &  u=0.
\end{array}\right.\label{6}
\end{equation}
\par Similar to M-estimators, a robust scale estimation can also be considered. However, improper setting of the scale may result in inability to convergence \cite{Huber1964Robust}. IRLS can be expressed by a sequence of successive minimizers of weighted $\ell_2$ based theoretical regularized risk, as follows:
\begin{equation}
f_{k+1}=\arg\min_{f} \ \tau||f||_{2}^2+E_{(X,Y)\sim P}[v(Y-f_{k}(X))(Y-f(X))^2].\label{3.1.2}
\end{equation}
\par To let the sequence $\{f_k\}$ converge, the following conditions have been proved to be necessary in \cite{irlskbr}:
\begin{description}
  \item[v1] $v(x)$ is a non-negative bounded Borel measurable function.
  \item[v2] $v(x)$ is an even function.
  \item[v3] $v(x)$ is continuous and differentiable, with $v'(x)\leq 0$ for $x>0$.
\end{description}
\par The sequence $\{f_k\}$ can be used as the solution of the $\ell$ based theoretical regularized risk. Specifically, the global minimum can be obtained as the limit of the sequence $\{f_k\}$ with arbitrary initialization if $\ell(\cdot)$ is convex. Otherwise, the $f_k$ $(k\to \infty)$, should be a local minimum depending on the initial start. Moreover, if $\ell(\cdot)$ is a robust loss function, the following requirements \cite{irlskbr} should be met:
\begin{description}
  \item[c1] $\psi(x)$ is a measurable, real, odd function.
  \item[c2] $\psi(x)$ is continuous and differentiable.
  \item[c3] $\psi(x)$ is bounded.
  \item[c4] $\psi(x)$ is increasing or strictly increasing.
\end{description}
\par However, in practical applications, the prior information of distribution $P$ is not available, that is, the expected risk can not be solved. Therefore, empirical risk is used to approximate the expected risk in Eq.(\ref{3.1.2}). Taking LS-SVR as an example, a sequence of successive minimizers of weighted LS-SVR can be expressed as:
\begin{equation}
(\bm{w}_{k+1},b_{k+1})=\arg\min_{\bm{w},b}\frac{1}{2} {\Vert \bm{w}\Vert}_2^2+\frac{C}{2}\sum_{i=1}^Nv(y_i-(\bm{w}_k^T\phi(\bm{x}_i)+b_k))
(y_i-(\bm{w}^T\phi(\bm{x}_i)+b))^2.
\label{214}
\end{equation}
\par The convergence and approximability of the sequence $\{\bm{w}_k, b_k\}$ will be proved latter. As described above, IRLS can be used to optimize any loss function based regressors, and it can also be used to interpret the robustness of arbitrary loss functions.
\section{Proposed methods}\label{sec:4}
\noindent In this section, we firstly introduce the proposed $\ell_s$-loss. Then two robust regressors are constructed on basis of $\ell_s$-loss, and optimized by IRLS technique. Finally, we prove the convergence of the proposed methods, and show the approximability between the utilized optimization technique and the original problem.
\subsection{The proposed $\ell_s$-loss function}\label{sec:4.0}
\noindent As stated before, the robustness of loss functions and regressors can be interpreted from the viewpoint of IRLS. In this subsection, a novel loss function, i.e., $\ell_s$-loss, is proposed for better noise robustness. Based on the observation that the sigmoid function $(1/(1+\exp(-u)))$ can be adjusted slightly to satisfy the requirement for the gradient function of a robust loss, i.e., \textbf{c1}-\textbf{c4}. Moreover, corresponding weight function meets the convergence conditions \textbf{v1}-\textbf{v3}. Therefore, a novel robust function can be obtained, which is induced from sigmoid function and defined as follows:
\begin{equation}
\ell_s(u)=ln(1+e^{\lambda u})-\frac{\lambda}{2}u+l_0
\label{4112}
\end{equation}
where $\lambda, l_0\in \mathbb{R}$. $\lambda$ is a hyperparameter used to control the amplitude, and $l_0$ is a constant to guarantee the loss function through the origin.\\ \\
\noindent \textbf{Remarks.} Properties of $\ell_s$-loss are highlighted as follows:
\begin{itemize}
  \item[-] $\ell_s$-loss is a convex, continuous and differentiable loss function, due to the smoothness and convexity, it can be optimized efficiently.
  \item[-] The gradient function of $\ell_s$-loss is a bounded, continuous, differentiable and strictly increasing odd function. Bounded gradient function theoretically leads to noise robustness.
  \item[-] The weight function of $\ell_s$-loss is a non-negative even function, and when its independent variable is greater than 0, its derivative is less than 0. This means that the samples with large residuals will be given smalle weights.
  \item[-] $\ell_s$-loss treats the samples with large residuals like $\ell_1$-loss, and both are more robust than $\ell_2$-loss, because when $u\to\infty$, the following limits hold:
  \begin{equation}
  \lim_{u\to \infty}\frac{\ell_s(u)}{\ell_1(u)}=\frac{\lambda}{2}, \ \ \ \lim_{u\to \infty}\frac{\ell_s(u)}{\ell_2(u)}=0.
  \end{equation} 
  \item[-] $\ell_s$-loss can be regarded as a generalized convex loss because of the asymptotic property between $\ell_1$ and $\ell_2$.
\end{itemize}
\begin{figure}[h]
	\begin{centering}
		\includegraphics[width=\textwidth]{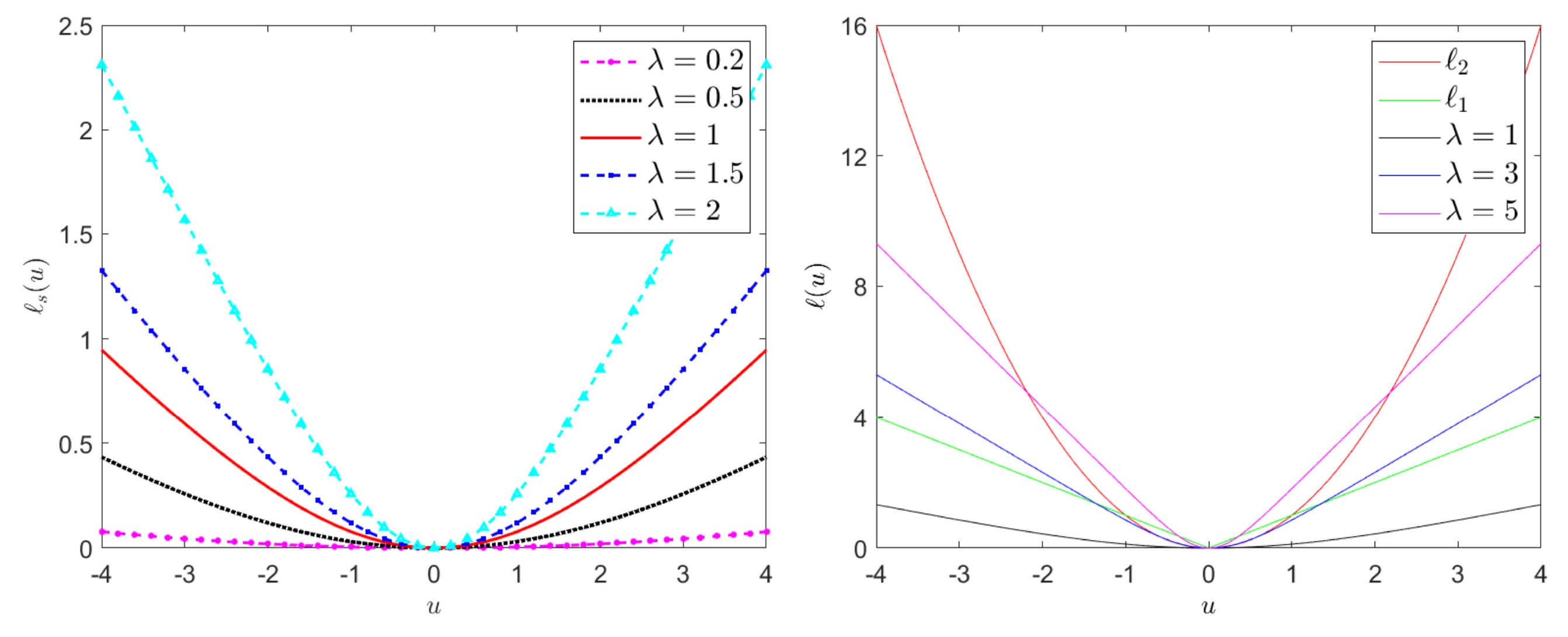}
		\caption{(Left) Form of the proposed $\ell_s$-loss under different values of hyperparameter $\lambda$. (Right) Comparison between  $\ell_2$, $\ell_1$ and $\ell_s$.}\label{fig2}
	\end{centering}
\end{figure}
\par As shown in Fig. \ref{fig2}, the height of $\ell_s$-loss will rise with the increase of $\lambda$. Moreover, the asymptotic property of $\ell_s$ between $\ell_1$ and $\ell_2$ can be seen from the right figure. When $\lambda=1$ (black solid line), the  shape of $\ell_s$-loss is relatively smooth, and the height is lower than $\ell_1$-loss. When $\lambda=3$ (blue solid line), $\ell_s$-loss approximate to $\ell_1$-loss, and when $\lambda=5$ (magenta solid line), $\ell_s$-loss is located between $\ell_1$-loss and $\ell_2$-loss. The larger value of $\lambda$, the sharper shape and the higher amplitude of the $\ell_s$-loss.
\subsection{Iteratively reweighted algorithm for $\ell_s$ based LS-SVR}
 \label{sec:4.1}
\noindent A robust kernel based regressor can be constructed by replacing the $\ell_2$-loss of LS-SVR with $\ell_s$-loss. Therefore, Eq.(\ref{3.1.1}) is changed to:
\begin{equation}
\min_{\bm{w},b} \ \ \ \frac{1}{2} {\Vert \bm{w}\Vert}_2^2+\frac{C}{2}\sum_{i=1}^N\ell_s(y_i-(\bm{w}^T\phi(\bm{x}_i)+b)).\label{4.1.1}
\end{equation}
\par As stated above, we solve Eq.(\ref{4.1.1}) by IRLS. Noting that the method based on $\ell_s$-loss and solved by IRLS technique is called IRLS-SVR in this paper. For a sequence of minimizers of $\ell_s$ based weighted LS-SVR, the $k+1$th iteration problem can be written as:
\begin{equation}
\min_{\bm{w},b} \ \ \  \frac{1}{2} {\Vert \bm{w}\Vert}_2^2+\frac{C}{2}\sum_{i=1}^N v_s(y_i-(\bm{w}_k^T\phi(\bm{x}_i)+b_k))
(y_i-(\bm{w}^T\phi(\bm{x}_i)+b))^2 \label{4}
\end{equation}
where $v_s(y_i-(\bm{w}_k^T\phi(\bm{x}_i)+b_k))$ represents the value of $\ell_s$-loss induced weight function which is computed by the $i$th sample and results of the $k$th iteration. Noting $v_s(y_i-(\bm{w}_k^T\phi(\bm{x}_i)+b_k))$ as $v_s(\bm{w}_k,b_k)$ and introducing the residual variable $\bm{\xi}$, Eq. (\ref{4}) can be rewritten as follows:
\begin{equation}
\begin{split}
(\bm{w}_{k+1}, b_{k+1})=\arg\min_{\bm{w},b} \ \ \  &\frac{1}{2} {\Vert \bm{w}\Vert}_2^2+\frac{C}{2}\sum_{i=1}^Nv_s(\bm{w}_{k}, b_{k})\xi_i^2 \\
s.t. \ \ \   &y_i=\bm{w}^T\phi(\bm{x}_i)+b+\xi_i, \ \ \ i=1,2,\cdots,N. \label{41}
\end{split}
\end{equation}
\par The Lagrangian function can be written as: 
\begin{equation}
\mathcal{L}(\bm{w},b,\bm{\xi},\bm{\alpha})=\frac{1}{2}{\Vert\bm{w}\Vert}_2^2+
\frac{C}{2}\sum_{i=1}^Nv_s(\bm{w}_{k}, b_{k})\xi_i^2-\sum_{i=1}^N\alpha_i(\bm{w}^T\phi(\bm{x}_i)
+b+\xi_i-y_i)\label{7}
\end{equation}
where $\bm{\alpha}>0$ is the Lagrangian multiplier. Setting the derivatives of Eq. (\ref{7}) to zero shows that:
\begin{equation}
\frac{\partial\mathcal{L}}{\partial \bm{w}}= \bm{w}-\sum_{i=1}^N\alpha_i \phi(\bm{x}_i)=0,\label{8}
\end{equation}
\begin{equation}
\frac{\partial\mathcal{L}}{\partial b}= -\sum_{i=1}^N\alpha_i =0,\label{888}
\end{equation}
\begin{equation}
\frac{\partial\mathcal{L}}{\partial \xi_i}=Cv_s(\bm{w}_{k}, b_{k})\xi_i-\alpha_i=0,\label{9}
\end{equation}
\begin{equation}
\frac{\partial\mathcal{L}}{\partial \alpha_i}=\bm{w}^T\phi(\bm{x}_i)+b+\xi_i-y_i=0.\label{10}
\end{equation}
\par Eliminating the variable $\bm{w}$ and $\bm{\xi}$, Eq. (\ref{8}) to Eq. (\ref{10}) can be transformed as:
\begin{equation}
\left[\begin{array}{cc}
    0 &    \bm{1^T}     \\
    \bm{1} &  K+V
\end{array}\right]
\left[\begin{array}{c}
    b      \\
    \bm{\alpha}
\end{array}\right]=
\left[\begin{array}{c}
    0     \\
    \bm{Y}
\end{array}\right]\label{11}
\end{equation}
where a weight matrix $V$ replaces the $E/C$ in Eq.(\ref{2}), which can be defined as:
\begin{equation}
V=\text{diag}(\frac{1}{Cv_{s,1}(\bm{w}_{k}, b_{k})},\frac{1}{Cv_{s,2}(\bm{w}_{k}, b_{k})},\cdots,\frac{1}{Cv_{s,N}(\bm{w}_{k}, b_{k})}),\label{4113}
\end{equation}
and initialized to the identity matrix. The convergence solution of Eq.(\ref{4}) will be used in Eq.(\ref{3}) for prediction. The iterative optimization process of IRLS-SVR is outlined as Algorithm \ref{alg:1}.
\begin{algorithm}[ht]
	\caption{Solving robust $\ell_s$-loss based LS-SVR using IRLS}
	\label{alg:1}
	\begin{algorithmic}[1]
		\STATE \textbf{Begin};
		\REQUIRE Training set $Z=\{\bm{x}_i,y_i\}_{i=1}^N$, $\bm{x}_i\in \mathbb{R}^{n}$, $y_i\in \mathbb{R}$.\\
		\ENSURE The optimal weight vector $\bm{\alpha}^*$ and bias $b^*$;\\
		\STATE Prepare Gaussian kernel matrix $K$, maximum number of iterations $M=100$, and iteration termination coefficient $\epsilon=10^{-4}$. Let $k=1$. \\
		\FOR{$k$ in $M$:}
		\STATE Calculate dual variable and bias $(\bm{\alpha}_{k}, b_k)$ by Eq. (\ref{11});\\
		\STATE Update residual variable $\bm{\xi}$ by $\bm{\xi}_{k}=Y-(K\bm{\alpha}_{k}+b_k)$;\\
		\STATE Update weight matrix $V$ by Eq. (\ref{4113}), and let $k=k+1$;\\
		\STATE If $||\bm{\alpha}_{k}-\bm{\alpha}_{k+1}||_2^2<\epsilon$, iteration process stop;\\
		\ENDFOR
		\RETURN $\bm{\alpha}^k$ and $b^k$
	\end{algorithmic}
\end{algorithm}
\subsection{Iteratively reweighted algorithm for $\ell_s$ based ELM}\label{sec:4.2}
\noindent In this part, the idea of IRLS-SVR is extended to ELM, and another robust method is constructed. Ordinary ELM \cite{Huang2006Extreme} can easily lead to overfitting, so the Tikhonov regularization is used as supplementary. A robust regularized ELM, i.e., IRLS-ELM, can be obtained by replacing the $\ell_2$ with $\ell_s$, as follows:
\begin{equation}
\min_{\bm{\beta}} \ \ \  \frac{1}{2} {\Vert \bm{\beta}\Vert}_2^2+\frac{C}{2}\sum_{i=1}^N\ell_s(y_i-\bm{h}(\bm{x}_i)\bm{\beta}).\label{321}
\end{equation}
\par Eq.(\ref{321}) can also be solved by IRLS. For a sequence of minimizers of $\ell_s$ based weighted ELM, the $k+1$th iteration
problem can be written as: 
\begin{equation}
\begin{split}
\bm{\beta}_{k+1}=\arg\min_{\bm{\beta}} \ \ \  &\frac{1}{2} {\Vert \bm{\beta}\Vert}_2^2+\frac{C}{2}\sum_{i=1}^Nv_s(\bm{\beta}_k)\xi_i^2\\
s.t. \ \ \   &y_i=\bm{h}(\bm{x}_i)\bm{\beta}+\xi_i, \ \ \ i=1,2,\cdots,N \label{322}
\end{split}
\end{equation}
where $v_s(\bm{\beta}_k)$ can be computed by $v_s(y_i-\bm{h}(\bm{x}_i)\bm{\beta}_k)$. By introducing the Lagrange multiplier $\bm{\alpha}$, the Lagrange function of Eq. (\ref{322}) can be expressed as:
\begin{equation}
\mathcal{L}(\bm{\beta},\bm{\xi},\bm{\alpha})=\frac{1}{2} {\Vert \bm{\beta}\Vert}_2^2+\frac{C}{2}\sum_{i=1}^N v_s(\bm{\beta}_k)\xi_i^2-\sum_{i=1}^N\alpha_i(\bm{h}(\bm{x}_i)\bm{\beta}+\xi_i-y_i).\label{325}
\end{equation}
\par Then setting the derivatives of Eq. (\ref{325}) to zero shows that:
\begin{equation}
\frac{\partial\mathcal{L}}{\partial \bm{\beta}}= \bm{\beta}^T-\sum_{i=1}^N\alpha_i \bm{h}(\bm{x}_i)=0,
\end{equation}
\begin{equation}
\frac{\partial\mathcal{L}}{\partial \xi_i}=Cv_s(\bm{\beta}_k)\xi_i-\alpha_i=0,
\end{equation}
\begin{equation}
\frac{\partial\mathcal{L}}{\partial \alpha_i}=\bm{h}(\bm{x}_i)\bm{\beta}+\xi_i-y_i=0.
\label{328}
\end{equation}
\par According to \cite{chen2017neurocom,4938676}, the solution of Eq. (\ref{322}) is
\begin{equation}
\bm{\beta_k}=\left\{
\begin{array}{cc}
    H^{T}(\frac{1}{C}+V HH^{T})^{-1}V Y, &    N<L     \\
     (\frac{1}{C}+H^{T} V H)^{-1}H^{T}V Y, &  N\geq L
\end{array}\right.\label{326}
\end{equation}
where the two forms of $\bm{\beta}$ are equivalent based on Woodbury identity, and $V$ is the weight matrix indunced by $\ell_s$-loss, which can be initialized to the identity matrix and defined as:
\begin{equation}
V=\text{diag}(v_{s,1}((\bm{\beta}_k),v_{s,2}((\bm{\beta}_k),\cdots,v_{s,N}((\bm{\beta}_k))\label{327}
\end{equation}
\par The convergence solution of Eq.(\ref{322}) will be used for prediction as:
\begin{equation}
f(\bm{x})=\bm{h}(\bm{x})\bm{\beta}^*.\label{327new}
\end{equation}
\par The iterative optimization process of IRLS-ELM is outlined as Algorithm \ref{alg:2}.\\
\begin{algorithm}[ht]
	\caption{Solving robust $\ell_s$-loss based ELM using IRLS}
	\label{alg:2}
	\begin{algorithmic}[1]
		\STATE \textbf{Begin};
		\REQUIRE Training set $Z=\{\bm{x}_i,y_i\}_{i=1}^N$, $\bm{x}_i\in \mathbb{R}^{n}$, $y_i\in \mathbb{R}$.\\
		\ENSURE The optimal weight of the output layer $\bm{\beta}^*$;\\
		\STATE Prepare hidden layer function $g(\bm{a}_i,b_i,\bm{x})$, number of hidden nodes $L$, maximum number of iterations $M=100$, and iteration termination coefficient $\epsilon=10^{-4}$. Let $k=1$. \\
		\FOR{$k$ in $M$:}
		\STATE Randomly generate parameters $(\bm{a}_i,b_i)$ of $L$ hidden nodes;\\
		\STATE Obtain the hidden layer output matrix $H$;\\
		\STATE Calculate optimization variable $\bm{\beta}_{k}$ by Eq. (\ref{326});\\
		\STATE Update residual variable $\bm{\xi}$ by $\bm{\xi}_{k}=Y-H\bm{\beta}_k$;\\
		\STATE Update weight matrix $V$ by Eq. (\ref{327}), and let $k=k+1$;\\
		\STATE If $||\bm{\beta}_{k}-\bm{\beta}_{k+1}||_2^2<\epsilon$, iteration process stop;\\
		\ENDFOR
		\RETURN $\bm{\beta}^k$
	\end{algorithmic}
\end{algorithm}
\subsection{Convergence and approximability}
\label{sec:4.3}
\noindent In this subsection, the convergence and approximability of the proposed methods are discussed. It is worthy noting that related conclusion was mentioned in \cite{irlskbr} on basis of the regularized theoretical risk. According to the statements of \cite{irlskbr}, for the $\ell_s$ based regularized theoretical risk, there exists a $f_\infty$ such that $f_k\to f_\infty$ as $k\to\infty$ since the $\ell_s$ induced weight function satisfies \textbf{v1}-\textbf{v3}. Besides, the convergence solution $f^{*}$ can approximate the optimal solution of the following objective:
\begin{equation}
\min_{f} \ \ \tau{\Vert f\Vert}_2^2+E_{(X,Y)\sim P}(\ell_s(Y-f(X))).
\label{431}
\end{equation}
\par We prove that this conclusion still holds when the regularized empirical risk of arbitrary convex loss fucntion $\ell(\cdot)$ is minimized. Taking LS-SVR with $\ell$-loss as an example, Eq.(\ref{431}) should be modified as:
\begin{equation}
\min_{\bm{w}} \ \ \tau {\Vert \bm{w}\Vert}_2^2+\frac{1}{N}
\sum_{i=1}^N\ell(y_i-\bm{w}^T\phi(\bm{x}_i))
\label{432}
\end{equation}
where the bias is deliberately omitted here for convenient. To be consistent, the $k+1$th segment of IRLS is given:
\begin{equation}
\bm{w}_{k+1}=\arg\min_{\bm{w}} \ \ \tau {\Vert \bm{w}\Vert}_2^2+\frac{1}{N}
\sum_{i=1}^Nv(y_i-\bm{w}_k^T\phi(\bm{x}_i))(y_i-\bm{w}^T\phi(\bm{x}_i))^2.
\label{433}
\end{equation}
\par Before the proof of convergence, the following representation theorem is given firstly:\\ \\
\textbf{Lemma 1.} \em Note the map $h(x,y)$ as $h_{k}(\bm{x},y)=y-\bm{w}_{k}^T\phi(\bm{x})$. For the result of $k+1$th iteration, it holds that:
\begin{equation}
\bm{w}_{k+1}=\frac{1}{\tau}\frac{1}{N}\sum_{i=1}^N[v(y_i-\bm{w}_k^T\phi(\bm{x}_i))
h_{k+1}(\bm{x}_i,y_i)\phi(\bm{x}_i)].
\label{434}
\end{equation}
\em\\ \\
\textbf{Proof.} For the optimization problem of Eq. (\ref{433}), Fermat Lemma gives the necessary conditions for the objective function to be extreme at some point. Due to $\bm{w}_{k+1}$ is the optimal solution of the $k$th iteration, the following equation holds:
\begin{equation}
\frac{\partial}{\partial \bm{w}_{k+1}} [\tau {\Vert \bm{w}_{k+1}\Vert}_2^2+\frac{1}{N}
\sum_{i=1}^Nv(y_i-\bm{w}_k^T\phi(\bm{x}_i))(y_i-\bm{w}_{k+1}^T\phi(\bm{x}_i))^2]=0.
\label{435}
\end{equation}
\par Expanding the derivative, we have:
\begin{equation}
2\tau\bm{w}_{k+1}-\frac{2}{N}
\sum_{i=1}^Nv(y_i-\bm{w}_k^T\phi(\bm{x}_i))(y_i-\bm{w}_{k+1}^T\phi(\bm{x}_i))\phi(\bm{x}_i)=0.
\end{equation}
\par The theorem can be proved by replacing $y_i-\bm{w}_{k+1}^T\phi(\bm{x}_i)$ with $h_{k+1}(\bm{x}_i,y_i)$.$\hfill\Box$ \\
\par The sequence $\{\bm{w}_k\}$ can be proved to be converge with the help of the above representation theorem. Noting the objective of Eq.(\ref{432}) as $R(\bm{w})$, it can be seen that $R\geq 0$ is obviously true. Therefore, we mainly focus on whether $\{R(\bm{w}_k)\}$ is strictly decreasing with the increase of $k$.\\ \\
\textbf{Theorem 1.} \em Initialize $w_0\in \mathbb{R}^m$ randomly. If the weight function $v(\cdot)$ induced by $\ell(\cdot)$-loss satisfying \textbf{v1}-\textbf{v3}, the solution of Eq.(\ref{433}) holds that $\bm{w}_k\to \bm{w}_\infty$, as $k\to\infty$.\em\\ \\
\textbf{Proof.} Define a real function $U(\cdot)$, which satisfies $U'(z)=\psi(z)=2zv(z)$. And a real function $g(\cdot)$, which satisfies $g(z^2)=U(z)$. We have $2zv(z)=(g(z^2))'=g'(z^2)\cdot 2z$, and $g'(z^2)=v(z)$. Because of \textbf{v1} and \textbf{v3}, it holds that $U'(z)\geq 0$ for $z\geq0$, and $U'(z)\leq 0$ for $z<0$. Due to \textbf{v3}, the weight function $v(\cdot)$ is decreasing, so the function $g(\cdot)$ is concave. So the inequality $g(a)-g(b)\leq(a-b)g'(b)$ holds. Therefore, the difference of objective function between two adjacent iterations can be written as:
\begin{equation}
\begin{aligned}
R(\bm{w}_{k+1})-R(\bm{w}_{k})&=\tau||\bm{w}_{k+1}||_2^2-\tau||\bm{w}_{k}||_2^2+\frac{1}{N}\sum_{i=1}^N[U(y_i-\bm{w}_{k+1}^T\phi(\bm{x}_i))-U(y_i-\bm{w}_{k}^T\phi(\bm{x}_i))] \\
&\leq\underbrace{\frac{1}{N}\sum_{i=1}^N[((y_i-\bm{w}_{k+1}^T\phi(\bm{x}_i))^2-(y_i-\bm{w}_{k}^T\phi(\bm{x}_i))^2)
g'((y_i-\bm{w}_{k}^T\phi(\bm{x}_i))^2)]}_{R_1}+\underbrace{\tau||\bm{w}_{k+1}||_2^2-\tau||\bm{w}_{k}||_2^2}_{R_2}.
\end{aligned}
\end{equation}
\par Based on $g'(z^2)=v(z)$ and difference of two squares, $R_1$ can be written as:
\begin{equation}
\frac{1}{N}\sum_{i=1}^N[v(y_i-\bm{w}_{k}^T\phi(\bm{x}_i))
(2y_i-\bm{w}_{k+1}^T\phi(\bm{x}_i)-\bm{w}_{k}^T\phi(\bm{x}_i))(\bm{w}_{k}^T\phi(\bm{x}_i)
-\bm{w}_{k+1}^T\phi(\bm{x}_i))].
\end{equation}
\par Substituting $(2y_i-2\bm{w}_{k+1}^T\phi(\bm{x}_i))+(\bm{w}_{k+1}^T\phi(\bm{x}_i)-\bm{w}_{k}^T\phi(\bm{x}_i))$ for $(2y_i-\bm{w}_{k+1}^T\phi(\bm{x}_i)-\bm{w}_{k}^T\phi(\bm{x}_i))$, and replacing $y_i-\bm{w}_{k+1}^T\phi(\bm{x}_i)$ with $h_{k+1}(\bm{x}_i,y_i)$, $R_1$ can be written in two parts, i.e., $R_{11}$ and $R_{12}$, as follows:
\begin{equation}
R_{11}=-\frac{1}{N}\sum_{i=1}^N[v(y_i-\bm{w}_{k}^T\phi(\bm{x}_i))
(\bm{w}_{k}^T\phi(\bm{x}_i)-\bm{w}_{k+1}^T\phi(\bm{x}_i))^2],\label{436}
\end{equation}
\begin{equation}
\begin{aligned}
R_{12}&=\frac{1}{N}\sum_{i=1}^N[v(y_i-\bm{w}_{k}^T\phi(\bm{x}_i))
(\bm{w}_{k}^T\phi(\bm{x}_i)-\bm{w}_{k+1}^T\phi(\bm{x}_i))2h_{k+1}(\bm{x}_i,y_i)]\\
&=(\bm{w}_{k}-\bm{w}_{k+1})^T\frac{1}{N}\sum_{i=1}^N[v(y_i-\bm{w}_{k}^T\phi(\bm{x}_i))
2h_{k+1}(\bm{x}_i,y_i)\phi(\bm{x}_i)].\label{4366}
\end{aligned}
\end{equation}
\par Using Lemma 1, Eq.(\ref{4366}) can be transformed as:
\begin{equation}
R_{12}=(\bm{w}_{k}-\bm{w}_{k+1})^T2\tau\bm{w}_{k+1}=-2\tau||\bm{w}_{k+1}||_2^2+2\tau\bm{w}_{k+1}^T\bm{w}_{k}.
\label{437}
\end{equation}
\par Therefore, the following formula holds:
\begin{equation}
R(\bm{w}_{k+1})-R(\bm{w}_{k})=R_{11}+R_{12}+R_2=R_{11}-\tau||\bm{w}_{k+1}||_2^2+2\tau\bm{w}_{k+1}^T\bm{w}_{k}
-\tau||\bm{w}_{k}||_2^2=R_{11}-\tau||\bm{w}_{k+1}-\bm{w}_{k}||_2^2
\end{equation}
\par It is obviously that $R_{11}$ is negative. So the set $\{R(\bm{w}_k)\}$ is strictly decreasing with the increase of iteration time $k$. Therefore, $\bm{w}_k\to \bm{w}_\infty$, as $k\to\infty$, and the theorem is proved.$\hfill\Box$ \\
\par It can be inferred from Theorem 1 that the proposed methods, i.e., IRLS-SVR and IRLS-ELM, can converge in theory. Next, we focus on the approximability of IRLS, i.e., the degree of approximation between the convergence solution of Eq.(\ref{432}) and the optimal solution of Eq.(\ref{431}). Without loss of generality, we still take LS-SVR as an example. The following proposition can be obtained:\\ \\
\noindent\textbf{Proposition 1.} \em The optimal solution of $\ell$ based LS-SVR can be obtained by solving a sequence of weighted LS-SVR with the weight function $v(\cdot)$ induced by $\ell$-loss under arbitrary initialization.\em \\ \\
\textbf{Proof.} For the set $\{\bm{w}\}$, which satisfies the requirements of convergence and obtained by solving a sequence of Eq.(\ref{432}), the limit $\bm{w}_{\infty}$ must meet the following condition based on Lemma 1,
\begin{equation}
\bm{w}_{\infty}=\frac{1}{\tau}\frac{1}{N}\sum_{i=1}^N[v(y_i-\bm{w}_{\infty}^T\phi(\bm{x}_i))
(y_i-\bm{w}_{\infty}^T\phi(\bm{x}_i))\phi(\bm{x}_i)].\label{439}
\end{equation}
\par A quantitative representation theorem for Eq. (\ref{432}) with arbitrary convex loss function was proposed in \cite{Christmann2008Support}, as follows:
\begin{equation}
\bm{w}=-\frac{1}{2\tau}\frac{1}{N}\sum_{i=1}^N\delta_i\phi(\bm{x}_i)\label{438}
\end{equation}
where $\delta_i=\ell'(y_i-\bm{w}^T\phi(\bm{x}_i))$, $\ell'$ denotes the derivative with respect to $\bm{x}$. Due to $\ell'(u)=\psi(u)=2uv(u)$, Eq. (\ref{438}) can be written as:
\begin{equation}
\bm{w}=-\frac{1}{2\tau}\frac{1}{N}\sum_{i=1}^N(\psi(y_i-\bm{w}^T\phi(\bm{x}_i))(-1)\phi(\bm{x}_i))=\frac{1}{2\tau}\frac{1}{N}\sum_{i=1}^N(v(y_i-\bm{w}^T\phi(\bm{x}_i))2(y_i-\bm{w}^T\phi(\bm{x}_i))\phi(\bm{x}_i)).\label{4310}
\end{equation}
\par Compare Eq. (\ref{439}) and Eq. (\ref{4310}), it can be found that the IRLS solution $\bm{w}_{\infty}$ satisfies the quantitative representation theorem for Eq. (\ref{432}). The proposition is proved due to the local optimal solution is certainly the global optimal one of convex optimization, and $\ell(\cdot)$ is a convex, continuous and differentiable loss function.$\hfill\Box$ \\
\par Without loss of generality, LS-SVR acts as an example in the proof, and these conclusions can be directly generalized to ELM. So far we theoretically prove the convergence and approximability of the proposed methods. Convergence is the basic condition of the algorithm. Approximability can help us interpret the robustness of the proposed methods.
\section{Robust analysis}
\label{sec:7}
\noindent In this section, we illustrate the robustness of the proposal. By directly observing the solution process of IRLS to the proposed methods, we can find that the samples which are difficult to predict will be gradually ignored as the iteration proceeds. 
The reason for this phenomenon is that the $\ell_s$ induced weight function $v_s(\cdot)$, which is defined as: 
\begin{equation}
v_s(u)=\frac{\lambda}{2u}(\frac{1}{1+\exp(-\lambda u)}-\frac{1}{2}),
\end{equation}
is a non-negative even function. Therefore, derivative of $v_s(\cdot)$ is negative, which means that for samples with large residuals, their weights will decrease with the increase of the residual and eventually tend to zero. 
\par In addition to the interpretion of robustness with the help of IRLS, it is also analyzed from theoretical and numerical aspects. Influence function and sensitivity curve are used as the tools of theoretical and numerical analysis, respectively. The results can reflect the robustness of the proposed methods to some extent.
\subsection{Theoretical perspective}
\label{sec:7.1}
\noindent As known, the derivative of a function describes its change rate at certain points. It is easy to conclude that the change rate of $\ell_2$-loss is very fast because its gradient is linear, which leads to a sharp form and noise sensitive. For robust loss functions, bounded derivatives are necessary \cite{Huber1964Robust}. From a theoretical point of view, bounded influence function (IF) means that the change of function value caused by noise has an upper limit \cite{Ref1006}. Influence function of estimator $T$ can be defined as:
\begin{equation}
IF=\lim_{\varepsilon\to 0}\frac{T((1-\varepsilon)F+\varepsilon H)-T(F)}{\varepsilon}
\end{equation}
where $F$ is the main distribution, $H$ is the pollution distribution and $\varepsilon$ is the pollution rate. Specifically, for loss function $\ell(\cdot)$, the above formula can be written as:
\begin{equation}
IF=M^{-1}\ell'(z_y-f(z_x))z_x
\end{equation}
where $z=(z_x,z_y)$ is the polluted point, $f$ is the prediction function, and $M$ is the mean of $\sum_{i}\ell''(y_i-f(x_i))x_i^Tx_i$.
\par Another related conclusion were presented in \cite{Christmann2007Consistency}, as follows:
\begin{equation}
IF=S^{-1}(E_F(\ell'(Y,f(X))\psi(X)))-\ell'(z_y,f(z_x))S^{-1}\psi(z_x)
\end{equation}
where $S(f)=4f/C+E_F((\ell''(Y,f(X))<\psi(X),f>\psi(X)))$. By comparing the above two formulas, it can be seen that bounded gradient and kernels (such as radial basis function, RBF) is the foundation of a bounded influence function. According to the representation of $\ell_s$-loss, its gradient function can be written as follows:
\begin{equation}
\psi_s(u)=\frac{\lambda}{1+\exp(-\lambda u)}-\frac{\lambda}{2}.\label{gradls}
\end{equation}
\par As stated before, the requirements of \textbf{c1}-\textbf{c4} should be met if it is a robust loss function. Therefore, one can easily prove that the proposed $\ell_s$-loss is a robust loss function from the expression of Eq. (\ref{gradls}).
\begin{figure}[ht]
	\begin{centering}
		\includegraphics[width=0.75\textwidth]{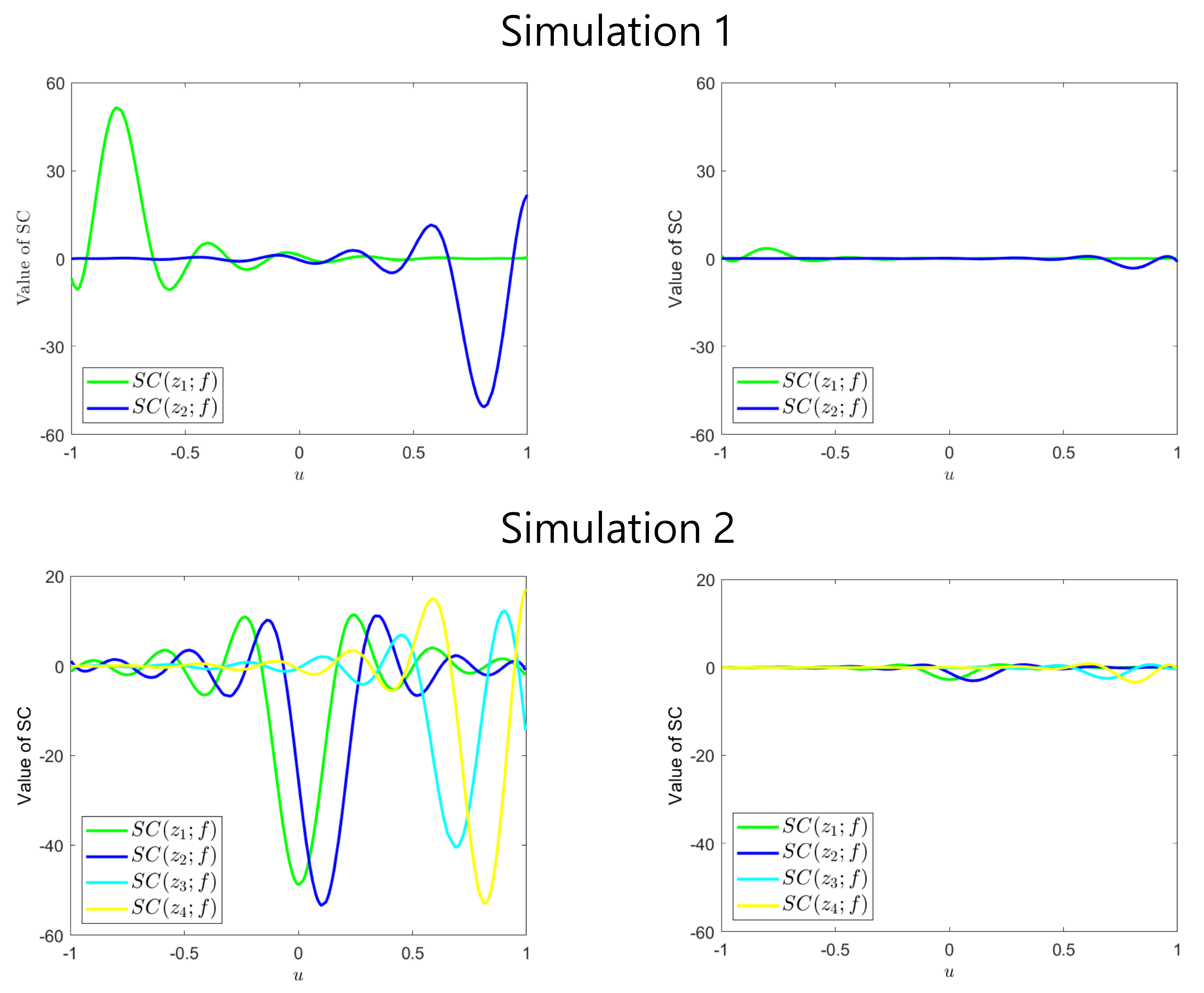}
		\caption{(Left) Sensitivity curves of LS-SVR. (Right) Sensitivity curves of the proposed IRLS-SVR. The training data of Simulation 1 is generated from $y=\sin(u)\cos(u^2)$, $u\in[-1,1]$, and $z_1=(-0.8,-5)$, $z_2=(0.8,5)$. Data of Simulation 2 is generated from $y=15(u^2-1)^2u^4\exp(-u)$, $u\in[-1,1]$, and $z_1=(0,5)$, $z_2=(0.1,5)$, $z_3=(0.7,5)$, $z_4=(0.8,5)$. Curves in different colors represent the SC of different outliers.}\label{fig_sc}
	\end{centering}
\end{figure}
\subsection{Numerical perspective}
\label{sec:7.2}
\noindent Even with the above analysis, it is still not easy to measure the robustness of the algorithm. In this part, we use sensitivity curve (SC) to show the robustness visually, which can be seen as finite version of the influence function. From \cite{irlskbr,Christmann2008Support,pinsvm}, SC at an additional point $z_i(x_i,y_i)$ can be defined as:
\begin{equation}
SC(z_i;f)=\frac{(f(P)-f(P^i))}{1/N}\label{sc}
\end{equation}
where $P$ is the training set with $N$ samples and $P^i=P\backslash z_i$. $f(P)$ and $f(P^i)$ represent the constructed decision functions with and without $z_i$, respectively. Generally, the additional point $z_i$ may be set as an errant point, i.e., outlier, so that we can see how it affects the model. It can be seen from the definition that SC reflects the impact of outliers  on decision-making, and a smaller value of SC undoubtedly means better robustness. 
\par Artificial data points with several deliberately added outliers are generated in two different settings. The first data set is obtained from $y=\sin(u)\cos(u^2)$, $u\in[-1,1]$, and two outliers are $z_1(-0.8,-5)$, $z_2(0.8,5)$. The second data set is obtained from $y=15(u^2-1)^2u^4\exp(-u)$, $u\in[-1,1]$, and four outliers are $z_1(0,5)$, $z_2(0.1,5)$, $z_3(0.7,5)$, $z_4(0.8,5)$. Using these data, 2D SC of LS-SVR and IRLS-SVR are drawn to show their robustness visually. According to Eq.(\ref{sc}), under different settings, the SC of each outlier is shown in Figure \ref{fig_sc}. It can be seen that the SC values of the proposal is significantly lower than the one of LS-SVR, which means that the robustness is actually improved.
\section{Experiments}
\label{sec:8}
\noindent Experiments on artificial and benchmark datasets are carried out in order to evaluate the effectiveness of proposed methods. Specifically, classical and advanced regressors are chosen for comparison. All the experiments are implemented on a personal computer with Intel(R) Core(TM) i5-3230M CPU with 4 GB RAM, and MATLAB R2014a environment. Root mean square error (RMSE), mean absolute error (MAE) and mean relative error (MRE) are selected as evaluation criterion.
\subsection{Simulation on synthetic data}
\label{sec:8.1}
\begin{figure}[ht]
	\begin{centering}
		\includegraphics[width=\textwidth]{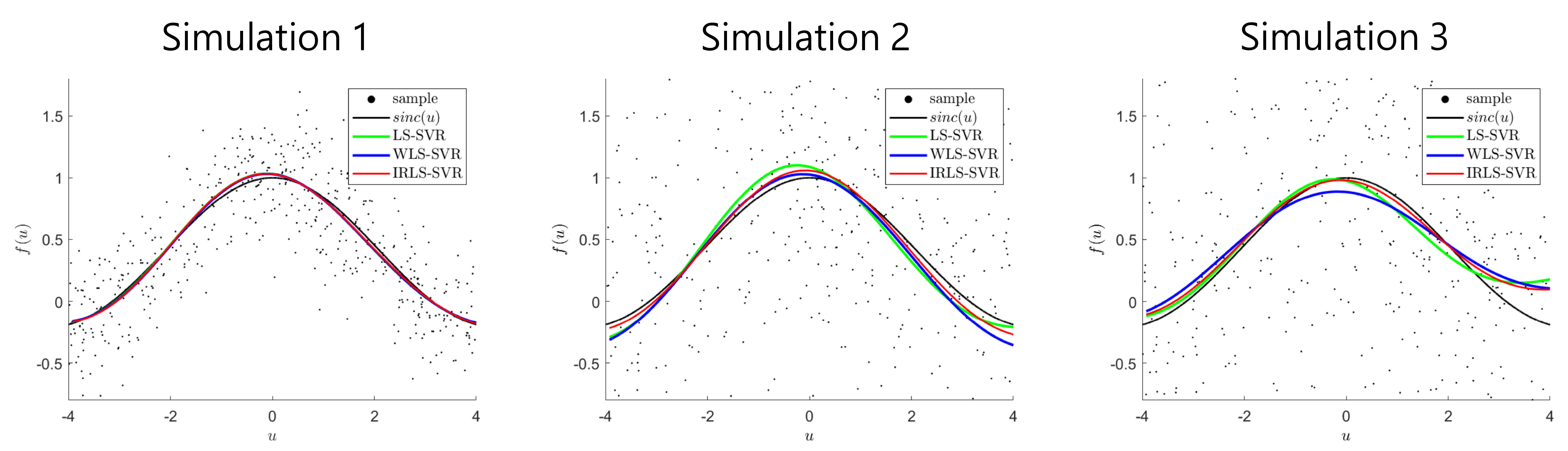}
		\caption{Regression curves of involved methods on artificial dataset. From left to right, Simulation 1-3 denote three kinds of noise, i.e., Gaussian $N(0,0.3^2)$, Laplacian $L(0,1)$ and $\chi^2(4)$, respectively. Curves in different colors represent different regressors.}\label{figrc}
	\end{centering}
\end{figure}

\noindent For synthetic data, we generate the training data from $sinc$ function. Moreover, noise subject to three kinds of  distributions, including Gaussian $N(0,0.3^2)$, Laplacian $L(0,1)$ and $\chi^2$ with 4 degree of freedom, is added to verify the robustness. IRLS-SVR is used in this part to compare with LS-SVR and weighted LS-SVR (WLS-SVR). There are totally $500$ training and $300$ testing samples, and the noise is only added to the training set. The process are repeated $500$ times to reduce randomness. Regression curves are drawn in Figure \ref{figrc}.
\par It can be obviously seen from Figure \ref{figrc} that the approximation degree of regression and real curves is relatively high under Gaussian noise. But $\ell_2$ based regressor and one-shot weighted are difficult to handle more complicated asymmetric distributed noises. The proposal has better performance because of the stronger capacity of $\ell_s$-loss as well as the IRLS solution. 
\begin{table}
\footnotesize
\caption{Experimental results of LS-SVRs on benchmark datasets without noise.}\label{table4}
\begin{tabular*}{\hsize}{@{}@{\extracolsep{\fill}}llllll@{}}
\midrule
Dataset &Method &$(C,\gamma,\lambda)$  &RMSE &MAE
&MRE   \\
\toprule
Diabetes  &LS-SVR  &$(2^7,2^{-1},\backslash)$ &\textbf{0.1493}     &0.1260    &0.3219   \\
$(43\times2)$&WLS-SVR &$(2^7,2^{-1},\backslash)$  &0.1497    &\textbf{0.1257}    &0.3217  \\
&IRLS-SVR   &$(2^{8},2^{-1},2^{1})$  &\textbf{0.1493}    &0.1260    &\textbf{0.3216}   \\
\midrule
Triazines  &LS-SVR  &$(2^5,2^{-3},\backslash)$ &\textbf{0.1636}    &\textbf{0.1226}    &\textbf{0.9146}    \\
$(186\times60)$&WLS-SVR &$(2^5,2^{-3},\backslash)$  &0.1730    &0.1246    &0.9411   \\
&IRLS-SVR   &$(2^{8},2^{-3},2^{0})$  &\textbf{0.1636}    &\textbf{0.1226}    &0.9148  \\
\midrule
Boston Housing	 &LS-SVR  &$(2^5,2^{-3},\backslash)$ &0.0870    &0.0642    &0.2350    \\
$(506\times14)$&WLS-SVR &$(2^5,2^{-3},\backslash)$  &0.0870    &\textbf{0.0629}    &\textbf{0.2270}   \\
&IRLS-SVR   &$(2^{2},2^{-3},2^{3})$  &\textbf{0.0858}    &0.0631    &0.2302   \\
\midrule
AutoMPG	&LS-SVR  &$(2^4,2^{0},\backslash)$ &0.0671    &0.0505    &0.1817    \\
$(392\times7)$&WLS-SVR &$(2^4,2^{0},\backslash)$     &\textbf{0.0662}    &0.0503    &\textbf{0.1774}   \\
&IRLS-SVR   &$(2^{1},2^{0},2^{3})$  & 0.0668    &\textbf{0.0502}   & 0.1804  \\
\midrule
Slumptest  &LS-SVR  &$(2^8,2^{-2},\backslash)$ &0.0239    &0.0190    &0.0692    \\
$(103\times10)$&WLS-SVR &$(2^8,2^{-2},\backslash)$  &0.0260    &0.0201    &0.0698   \\
&IRLS-SVR   &$(2^{8},2^{-2},2^{3})$  & \textbf{0.0194 }   &\textbf{0.0149}   &\textbf{0.0610}  \\
\midrule
MachineCPU	&LS-SVR  &$(2^6,2^{-3},\backslash)$ &0.0488   & 0.0294    &0.5912    \\
$(209\times7)$&WLS-SVR &$(2^6,2^{-3},\backslash)$  &0.0533    &0.0315    &0.6432   \\
&IRLS-SVR   &$(2^{3},2^{-3},2^{3})$  &\textbf{ 0.0486}    &\textbf{0.0293 }   &\textbf{0.5909 } \\
\midrule
\end{tabular*}
\end{table}
\begin{table}
\footnotesize
		\caption{Experimental results of ELMs on benchmark datasets without noise.}\label{table5}
		\begin{tabular*}{\hsize}{@{}@{\extracolsep{\fill}}llllll@{}}
			\midrule
			Dataset &Method &$(C,L,\lambda)$  &RMSE &MAE
			&MRE   \\
			\toprule
			Diabetes  &ELM  &$(2^6,8,\backslash)$ &$0.1620\pm0.0129$&$0.1346\pm0.0091$&$0.3618\pm0.0355$\\
			&W-ELM  &$(2^6,8,\backslash)$  &$\textbf{0.1558}\pm\textbf{0.0041}$&$\textbf{0.1306}\pm\textbf{0.0020}$&$\textbf{0.3309}\pm\textbf{0.0135}$\\
			&IRLS-ELM   &$(2^{-4},8,2^{1})$  &$0.1770\pm0.0185$&$0.1444\pm0.0095$&$0.3762\pm0.0177$\\
			\midrule			
			Triazines  &ELM   &$(2^6,18,\backslash)$ &$0.1887\pm0.0071$&$0.1440\pm0.0044$&$1.0280\pm0.0597$\\
			&W-ELM   &$(2^6,18,\backslash)$  &$0.1871\pm0.0052$&$\textbf{0.1379}\pm\textbf{0.0045}$&$1.0370\pm0.1340$\\
			&IRLS-ELM   &$(2^{1},18,2^{2})$  &$\textbf{0.1867}\pm\textbf{0.0058}$&$0.1410\pm0.0050$&$\textbf{0.9628}\pm\textbf{0.0615}$\\
			\midrule
			Boston Housing	 &ELM  &$(2^8,25,\backslash)$ &$0.1114\pm0.0053$&$0.0832\pm0.0036$&$0.3135\pm0.0126$\\
			&W-ELM  &$(2^8,25,\backslash)$ &$0.1125\pm0.0023$&$0.0843\pm0.0019$&$0.3349\pm0.0109$\\
			&IRLS-ELM   &$(2^{0},25,2^{3})$  &$\textbf{0.1077}\pm\textbf{0.0010}$&$\textbf{0.0810}\pm\textbf{0.0016}$&$\textbf{0.2973}\pm\textbf{0.0169}$\\
			\midrule
			AutoMPG	&ELM  &$(2^3,39,\backslash)$ &$\textbf{0.0743}\pm\textbf{0.0021}$&$\textbf{0.0565}\pm\textbf{0.0014}$&$\textbf{0.2039}\pm\textbf{0.0138}$\\
			&W-ELM  &$(2^3,39,\backslash)$   &$0.0753\pm0.0019$&$0.0573\pm0.0013$&$0.2160\pm0.0129$\\
			&IRLS-ELM   &$(2^{7},39,2^{3})$  &$0.0760\pm0.0030$&$0.0578\pm0.0024$&$0.2141\pm0.0230$\\
			\midrule
			Slumptest &ELM    &$(2^{-1},51,\backslash)$ &$0.0598\pm0.0084$&$0.0475\pm0.0077$&$0.1804\pm0.0371$\\
			&W-ELM    &$(2^{-1},51,\backslash)$ &$0.0634\pm0.0195$&$0.0481\pm0.0128$&$0.1757\pm0.0523$\\
			&IRLS-ELM   &$(2^{-4},51,2^{1})$  &$\textbf{0.0569}\pm\textbf{0.0122}$&$\textbf{0.0437}\pm\textbf{0.0084}$&$\textbf{0.1473}\pm\textbf{0.0318}$\\
			\midrule
			MachineCPU	&ELM  &$(2^{-1},10,\backslash)$ &$0.0633\pm0.0049$&$0.0395\pm0.0023$&$0.9238\pm0.0873$\\
			&W-ELM  &$(2^{-1},10,\backslash)$   &$0.0637\pm0.0094$&$\textbf{0.0375}\pm\textbf{0.0043}$&$\textbf{0.8256}\pm\textbf{0.1219}$\\
			&IRLS-ELM   &$(2^{3},10,2^{2})$  &$\textbf{0.0612}\pm\textbf{0.0049}$&$0.0385\pm0.0034$&$0.9386\pm0.0761$\\
			\midrule
		\end{tabular*}
\end{table}
\begin{table}
	\footnotesize
	\caption{Experimental results of LS-SVRs on benchmark datasets with noise.}\label{table4.1}
	\begin{tabular*}{\hsize}{@{}@{\extracolsep{\fill}}llllll@{}}
		\midrule
		Dataset &Method &$(C,\gamma,\lambda)$  &RMSE &MAE
		&MRE   \\
		\toprule
		Diabetes  &LS-SVR  &$(2^7,2^{-1},\backslash)$&$1.0575\pm0.1329$&$0.9345\pm0.1354$&$2.3504\pm0.3941$\\
		&WLS-SVR &$(2^7,2^{-1},\backslash)$  &$0.4451\pm0.1618$ &$0.3555\pm0.1355$ &$0.9990\pm0.4146$\\
		&IRLS-SVR   &$(2^{8},2^{-1},2^{1})$  &$\textbf{0.4255}\pm\textbf{0.0860}$&$\textbf{0.3497}\pm\textbf{0.0672}$&$\textbf{0.9942}\pm\textbf{0.2268}$\\
		\midrule		
		Triazines  &LS-SVR  &$(2^5,2^{-3},\backslash)$ &$1.9586\pm0.1943$&$1.4591\pm0.1460$&$5.4298\pm1.6316$\\
		&WLS-SVR &$(2^5,2^{-3},\backslash)$  &$\textbf{1.4187}\pm\textbf{0.2357}$&$\textbf{0.8559}\pm\textbf{0.1321}$&$\textbf{2.5385}\pm\textbf{0.8062}$\\
		&IRLS-SVR   &$(2^{8},2^{-3},2^{0})$  &$1.6620\pm0.2024$&$1.1235\pm0.1169$&$3.0085\pm0.5551$\\
		\midrule
		Boston Housing	 &LS-SVR  &$(2^5,2^{-3},\backslash)$ &$0.8322\pm0.0738$&$0.7147\pm0.0740$&$2.1248\pm0.2220$\\
		&WLS-SVR &$(2^5,2^{-3},\backslash)$ &$0.2133\pm0.0331$&$0.1639\pm0.0254$&$0.7627\pm0.1705$\\
		&IRLS-SVR   &$(2^{2},2^{-3},2^{3})$  &$\textbf{0.1339}\pm\textbf{0.0130}$&$\textbf{0.1095}\pm\textbf{0.0114}$&$\textbf{0.4614}\pm\textbf{0.0730}$\\
		\midrule
		AutoMPG	&LS-SVR  &$(2^4,2^{0},\backslash)$ &$0.9354\pm0.0393$&$0.7451\pm0.0525$&$2.1356\pm0.1314$\\
		&WLS-SVR &$(2^4,2^{0},\backslash)$   &$0.2842\pm0.0473$&$0.1946\pm0.0275$&$0.8632\pm0.0983$\\
		&IRLS-SVR   &$(2^{1},2^{0},2^{3})$  &$\textbf{0.1421}\pm\textbf{0.0100}$&$\textbf{0.1092}\pm\textbf{0.0086}$&$\textbf{0.4560}\pm\textbf{0.0224}$\\
		\midrule
		Slumptest  &LS-SVR  &$(2^8,2^{-2},\backslash)$ &$1.6551\pm0.2270$&$1.3237\pm0.1675$&$4.3510\pm0.9127$\\
		&WLS-SVR &$(2^8,2^{-2},\backslash)$ &$\textbf{1.2711}\pm\textbf{0.3077}$&$\textbf{1.0014}\pm\textbf{0.2654}$&$\textbf{3.1439}\pm\textbf{0.5351}$\\
		&IRLS-SVR   &$(2^{8},2^{-2},2^{3})$  &$1.3975\pm0.1901$&$1.1127\pm0.1646$&$3.5497\pm0.5067$\\
		\midrule
		MachineCPU	&LS-SVR &$(2^6,2^{-3},\backslash)$ &$0.2697\pm0.0499$&$0.1862\pm0.0315$&$3.4087\pm0.5773$\\
		&WLS-SVR &$(2^6,2^{-3},\backslash)$  &$0.0872\pm0.0048$&$\textbf{0.0613}\pm\textbf{0.0044}$&$\textbf{1.6631}\pm\textbf{0.0629}$\\
		&IRLS-SVR   &$(2^{3},2^{-3},2^{3})$  &$\textbf{0.0851}\pm\textbf{0.0142}$&$0.0643\pm0.0090$&$1.8202\pm0.1500$\\
		\midrule
	\end{tabular*}
\end{table}
\begin{table}
	\footnotesize
	\caption{Experimental results of ELMs on benchmark datasets with noise.}\label{table5.1}
	\begin{tabular*}{\hsize}{@{}@{\extracolsep{\fill}}llllll@{}}
		\midrule
		Dataset &Method &$(C,L,\lambda)$  &RMSE &MAE
		&MRE   \\
		\toprule
		Diabetes  &ELM  &$(2^6,8,\backslash)$ &$1.1969\pm0.1523$&$1.0704\pm0.1222$&$2.6609\pm0.5126$\\
		&W-ELM  &$(2^6,8,\backslash)$  &$0.5886\pm0.0787$&$0.4793\pm0.0693$&$1.3178\pm0.2761$\\
		&IRLS-ELM   &$(2^{-4},8,2^{1})$  &$\textbf{0.4753}\pm\textbf{0.0861}$&$\textbf{0.3955}\pm\textbf{0.0725}$&$\textbf{1.1265}\pm\textbf{0.1497}$\\
		\midrule
		Triazines  &ELM   &$(2^6,18,\backslash)$ &$1.5228\pm0.1205$&$1.3004\pm0.0858$&$4.1854\pm0.9463$\\
		&W-ELM   &$(2^6,18,\backslash)$  &$0.6436\pm0.1223$&$0.4613\pm0.0776$&$2.0766\pm0.5883$\\
		&IRLS-ELM   &$(2^{1},18,2^{2})$  &$\textbf{0.2984}\pm\textbf{0.0083}$&$\textbf{0.2259}\pm\textbf{0.0050}$&$\textbf{1.4069}\pm\textbf{0.2056}$ \\
		\midrule
		Boston Housing	 &ELM  &$(2^8,25,\backslash)$ &$0.1114\pm0.0053$&$0.0832\pm0.0036$&$0.3135\pm0.0126$\\
		&W-ELM  &$(2^8,25,\backslash)$ &$0.1125\pm0.0023$&$0.0843\pm0.0019$&$0.3349\pm0.0109$\\
		&IRLS-ELM   &$(2^{0},25,2^{3})$  &$\textbf{0.1077}\pm\textbf{0.0010}$&$\textbf{0.0810}\pm\textbf{0.0016}$&$\textbf{0.2973}\pm\textbf{0.0169}$\\
		\midrule
		AutoMPG	&ELM  &$(2^3,39,\backslash)$ &$\textbf{0.0743}\pm\textbf{0.0021}$&$\textbf{0.0565}\pm\textbf{0.0014}$&$\textbf{0.2039}\pm\textbf{0.0138}$\\
		&W-ELM  &$(2^3,39,\backslash)$   &$0.0753\pm0.0019$&$0.0573\pm0.0013$&$0.2160\pm0.0129$\\
		&IRLS-ELM   &$(2^{7},39,2^{3})$  &$0.0760\pm0.0030$&$0.0578\pm0.0024$&$0.2141\pm0.0230$\\
		\midrule
		Slumptest &ELM    &$(2^{-1},51,\backslash)$ &$2.5533\pm0.2901$&$2.0344\pm0.2063$&$8.1882\pm1.3991$\\
		&W-ELM    &$(2^{-1},51,\backslash)$ &$2.8802\pm0.5165$&$2.2698\pm0.4157$&$7.9072\pm1.6204$\\
		&IRLS-ELM   &$(2^{-4},51,2^{1})$  &$\textbf{2.5214}\pm\textbf{0.1040}$&$\textbf{1.9153}\pm\textbf{0.0608}$&$\textbf{7.0224}\pm\textbf{0.5167}$ \\
		\midrule
		MachineCPU	&ELM  &$(2^{-1},10,\backslash)$ &$0.3381\pm0.0633$&$0.2233\pm0.0229$&$5.8377\pm1.3531$\\
		&W-ELM  &$(2^{-1},10,\backslash)$   &$0.1305\pm0.0183$&$\textbf{0.0866}\pm\textbf{0.0062}$&$\textbf{2.0914}\pm\textbf{0.2821}$\\
		&IRLS-ELM   &$(2^{3},10,2^{2})$  &$\textbf{0.1262}\pm\textbf{0.0176}$&$0.0959\pm0.0090$&$2.8142\pm0.4807$ \\
		\midrule
	\end{tabular*}
\end{table}
\subsection{Simulation on benchmark data}
\label{sec:8.2}
\noindent We test the proposed methods with several classical regressors on six UCI benchmark datasets \cite{Blake1998UCI} to illustrate their effectiveness. For IRLS-SVR, LS-SVR \cite{Suykens2002Least} and weighted LS-SVR (WLS-SVR) \cite{Suykens2002Weighted} are chosen for comparison. ELM \cite{Huang2006Extreme} and weighted ELM (W-ELM) \cite{4938676} are used to compare with IRLS-ELM. Information about the used datasets can be found from Table \ref{table4}. All samples are scaled to let the features locate in $[0,1]$ so as to improve the performance. Testing accuracy is obtained by using 10-fold cross validation \cite{Ref1010}. For LS-SVR and its variants, the hyperparameter $C$ is selected from $\{2^i|i=-4,\cdots,8\}$, and the rest ones are selected from $\{2^i|i=-3,\cdots,3\}$ by grid search. For ELMs, the selection of $C,\lambda$ is the same as that of LS-SVR, and the optimal number of hidden nodes $L$ is chosen from $\{r\cdot N|r=5\%,10\%,20\%,\cdots,50\%\}$ where $N$ is the number of training samples.  Noise-free and noisy experiments are carried out. For the noisy experiment, $20\%$ training samples are randomly selected to simulate outliers by amplifying their regression values ten times. Besides, experiments are implemented five times and the mean and variance are recorded in order to reduce the randomness. Noise-free experimental results are summarized in Table \ref{table4} for LS-SVRs and Table \ref{table5} for ELMs.
\par From Tables \ref{table4} and \ref{table5}, one can see that when training without outliers, the accuracy of the proposal is comparable to that of comparing methods. As shown in Tables \ref{table4.1} and \ref{table5.1}, when the training set is contaminated by outliers, results of $\ell_2$ based methods are unsatisfactory in all datasets which reflects the sensitivity of $\ell_2$-loss to outliers. The difference between multiple and one-shot weighted processes are the design of weight functions and the frequency of weighted operations. These two points results in the difference in their performance. Compared with $\ell_2$ based methods, the improved ones have more superior performance, and the proposed IRLS-SVR and IRLS-ELM achieve better results in more datasets. It is worhty noting that the value of hyperparameters is not re-selected for noisy experiment, but the value obtained from the previous noise-free experiment is reused. Therefore, such results are acceptable and sufficient to verify the robustness of the proposed methods.
\begin{figure}[ht]
	\begin{centering}
		\includegraphics[width=0.85\textwidth]{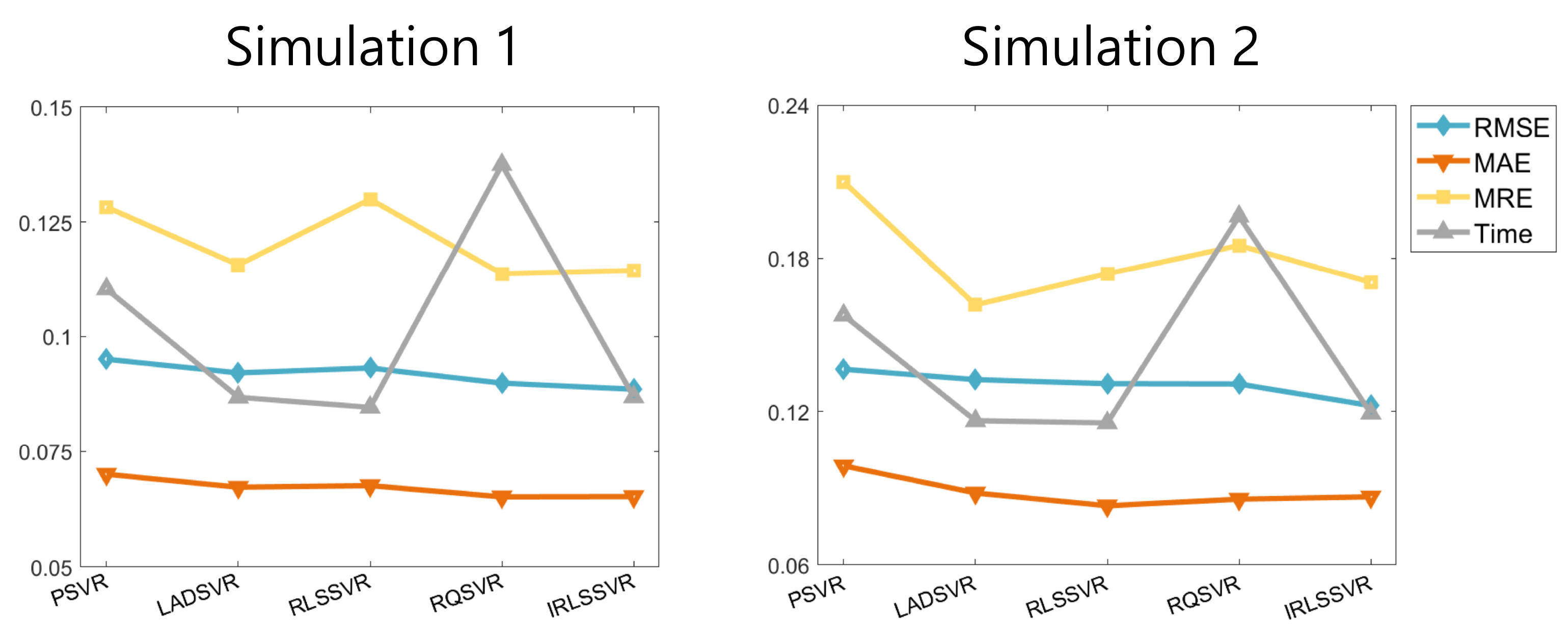}
		\caption{Experimental comparisons between several advanced methods and proposed method with noise. (Left) Results on AutoMPG dataset. (Right) Results on Boston Housing dataset. Values of MRE and running time are scaled for better presentation.}\label{fig6}
	\end{centering}
\end{figure}
\subsection{Comparison with advanced methods}
\label{sec:8.add}
\noindent In order to further verify the performance of the proposed $\ell_s$-loss, various advanced LS-SVR based methods, including pinball loss SVR (PSVR) \cite{Crambes2013}, least absolute deviation SVR (LADSVR) \cite{Chen2017Least}, robust non-convex LS-SVR (RLSSVR) \cite{Wang2014Robust}, robust generalized quantile loss based SVR (RQSVR) \cite{Yang2020} are selected to implement comparative experiments. We select two benchmark datasets, i.e., AutoMPG and Boston Housing, for verification. The experimental results are drawn in Figure \ref{fig6}.
\par In the experiment, we randomly select $5\%$ of the labeled samples from each dataset to choose the optimal hyperparameters. Then $20\%$ labeled samples are used for training and the rest for testing. Similar to before, artificial outliers are added to the training set. The whole process is repeated $50$ times to reduce the randomness. As shown in Figure \ref{fig6}, the experiments on two datasets show the same conclusion that the proposed method not only has good prediction accuracy, but also has good speed. It is worth noting that compared with the comparing methods, the proposal is the easiest to be implemented. It doesn't need to solve the time-consuming quadratic programming problem or involve advanced optimization methods, such as CCCP for RLSSVR and Split-Bregman iteration for LADSVR.
\par From the above experimental results including LS-SVR and ELM based regressions, we make the following summary:
\begin{itemize}
	\item[-] In noise-free experiments, performance of the proposal is close to comparing methods. The reason for such a phenomenon is that the goal of this work is to improve the robustness, and the dataset polluted by noise is the hypothetical application scene. 
	\item[-] Compared with classical methods, the proposed methods have better performance in most cases although the hyperparameters are not optimized. This shows that the proposed $\ell_s$-loss can improve the noise robustness of baselines in a plug-and-play manner.	
	\item[-] Compared with advanced kernel based regressors, the proposal has at least comparable robustness. We believe that the robustness comes from the proposed $\ell_s$-loss, which meets the theoretical requirements of robust loss function. Besides, the running speed of the proposed methods is satisfactory, which is closely related to the good properties of $\ell_s$-loss and the efficient solution brought by IRLS.
\end{itemize}
\subsection{Effect of hyperparameters}
\label{sec:8.3}
\begin{figure}
	\begin{centering}
		\includegraphics[width=0.9\textwidth]{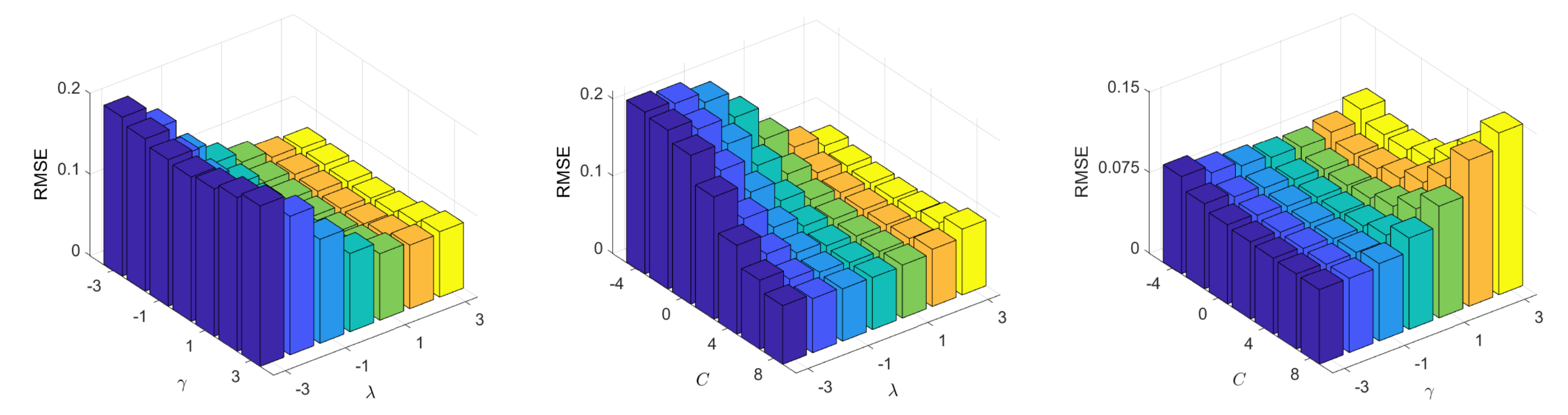}
		\caption{Effect of hyperparameters on RMSE of IRLS-SVR. From left to right, histograms show the influence of any two of the three on the model.}\label{fig7.1}
	\end{centering}
\end{figure}

\begin{figure}
	\begin{centering}
		\includegraphics[width=0.9\textwidth]{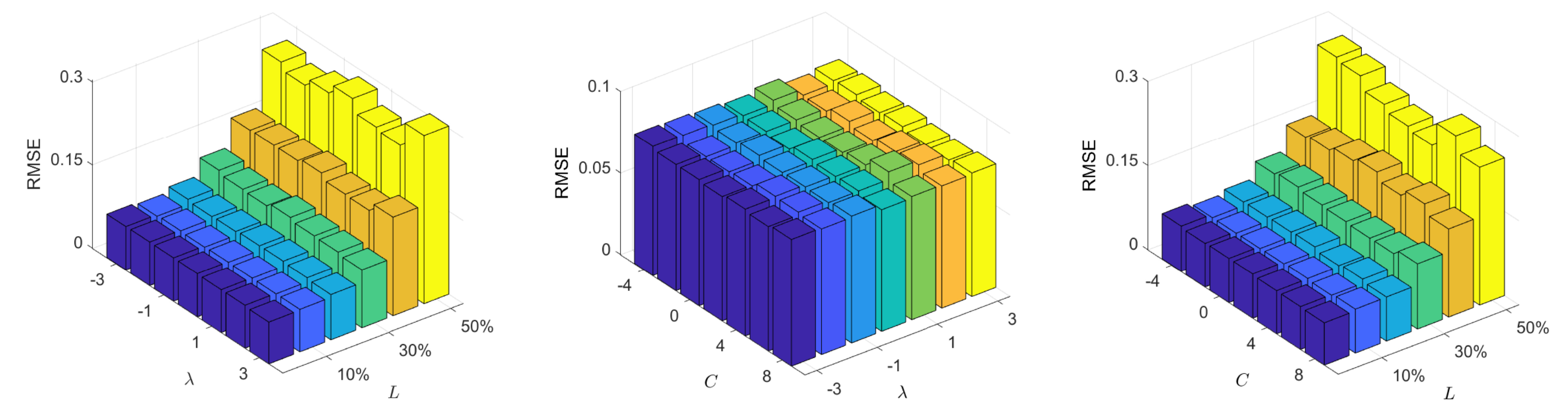}
		\caption{Effect of hyperparameters on RMSE of IRLS-ELM. From left to right, histograms show the influence of any two of the three on the model.}\label{fig7.2}
	\end{centering}
\end{figure}

\noindent In this part, experiments are carried out to reveal the influence of hyperparameters on the proposed methods. Specifically, AutoMPG dataset is used to obtain the RMSE under different hyperparameter combinations. Results of IRLS-SVR are shown in Figure \ref{fig7.1} as 3D histograms, and results of IRLS-ELM can be seen from Figure \ref{fig7.2}.
\par Figure \ref{fig7.1} shows the experimental results of the proposed IRLS-SVR. It can be seen from the figure that each hyperparameter has a certain effect on the model. When $C$ is fixed, the change of kernel parameter $\gamma$ has little effect on the model, while a larger value of $\lambda$ will make the model performance better. When $\gamma$ is fixed, the larger values of  $C$ and $\lambda$ are more efficient. When $\lambda$ is fixed, a medium size of $C$ and a small value of $\gamma$ are slightly better. In summary, the chosen of $\lambda$ should be larger, and $\gamma$ should be smaller, which will make the model performs better.
\par Figure \ref{fig7.2} shows the experimental results of the proposed IRLS-ELM. It can be concluded that the number of hidden nodes $L$ has a great impact on the performance, and the influence of the remaining ones is relatively small. Specifically, when the value of $C$ is fixed, the change in $\lambda$ has little effect on RMSE, but a smaller $L$ will significantly enhance the performance. Similarly, the same phenomena can be observed when fixing the value of $\lambda$. Besides, the value of $C$ and $\lambda$ have little effect when the optimal value of $L$ is selected. Therefore, more attention should be paid to search for a applicable number of hidden nodes when determining the value of hyperparameters. However, the number of $L$ is not as large as possible because more hidden nodes will reduce the calculation speed.
\section{Conclusion}
\label{sec:6}
\noindent In this paper, we propose a robust $\ell_s$ loss function, which is continuously and derivable, and its gradient function is bounded and strictly increasing. Simultaneously, devoting the proposed $\ell_s$-loss into least squares kernel based regressors and relacing the $\ell_2$-loss, two models, i.e., IRLS-SVR and IRLS-ELM, are constructed for better noise robustness. Moreover, IRLS technique is utilized to optimize and interpret the proposed methods. This work starts with the improvement of $\ell_2$ loss function, and the effectiveness can be observed by analyzing the weighted process of the model solution. In addition, convergence of the proposal has been proved theoretically. Global optimum can be achieved due to the convexity of the proposed $\ell_s$ loss function. Thanks to the novel loss function, the proposed methods achieve good results in the experiments on both artificial and benchmark datasets. The effect of hyperparameters is also discussed. 
\par Although achieves promising results, the difficulty of hyperparameter selection will result in huge training costs. In the future work, we will consider how to combine the proposed methods with hyperparameter optimization to address this issue.\\  \\
\noindent {\bf{Acknowledgements}}\\
This work is supported by National Nature Science Foundation of China (Nos. 11471010, 11271367).
{}


\begin{thebibliography}{}
\bibitem{Deep}
Krizhevsky A, Sutskever I, Hinton GE (2012) ImageNet classification with deep convolutional neural networks.
In: Proceedings of Advances in Neural Information Processing Systems, pp 1097--1105

\bibitem{Lecun2014Backpropagation}
Lecun Y, Boser B, Denker JS, Henderson D, Howard RE, Hubbard W, Jackel LD (1989) Backpropagation applied to handwritten zip code recognition. Neural Comput 1(4):541--551
\newblock \href {http://dx.doi.org/10.1162/neco.1989.1.4.541}
{\path{doi:10.1162/neco.1989.1.4.541}}

\bibitem{Audibert2011Robust}
Audibert JY, Catoni O (2011) Robust linear least squares regression. Ann Stat 39(5):2766--2794
\newblock \href {http://dx.doi.org/10.1214/11-AOS918}
{\path{doi:10.1214/11-AOS918}}

\bibitem{add4}
Cheung YM, Zeng H (2009) Local kernel regression score for selecting features of high-dimensional data. IEEE Trans Knowl Data Eng 21(12):1798-1802 
\newblock \href {http://dx.doi.org/10.1109/TKDE.2009.23}
{\path{doi:10.1109/TKDE.2009.23}}

\bibitem{Suykens2002Least}
Suykens JAK, Gestel TV, Brabanter JD, Moor BD, Vandewalle J (2002) Least squares support vector machines. Int J Circuit Theory Appl 27(6):605--615
\newblock \href{http://dx.doi.org/10.1002/(SICI)1097-007X(199911/12)27:6<605::AID-CTA86>3.0.CO;2-Z}
{\path{doi:10.1002/(SICI)1097-007X(199911/12)27:6<605::AID-CTA86>3.0.CO;2-Z}}

\bibitem{Huang2006Extreme}
Huang GB, Zhu QY, Siew CK (2006) Extreme learning machine: Theory and
applications. Neurocomputing 70(1):489--501
\newblock \href {http://dx.doi.org/10.1016/j.neucom.2005.12.126}
{\path{doi:10.1016/j.neucom.2005.12.126}}

\bibitem{vapniksvm}
Vapnik VN (2008) Statistical learning theory. Wiley, New York

\bibitem{Ref16}
Bartlett P, Mendelson S (2006) Empirical minimization. Probab Theory Relat Field 135(3):311--334
\newblock \href {http://dx.doi.org/10.1007/s00440-005-0462-3}
{\path{doi:10.1007/s00440-005-0462-3}}

\bibitem{Ref17}
Fama F, MacBeth D, Jackel LD (1973) Risk, return, and equilibrium: Empirical
tests. J Polit Econ 81(3):607--636
\newblock \href {http://dx.doi.org/10.1086/260061} {\path{doi:10.1086/260061}}


\bibitem{Catoni2010Challenging}
Catoni O (2010) Challenging the empirical mean and empirical variance: A deviation
study. Ann Inst Henri Poincare-Probab Stat 48(4):1148--1185
\newblock \href {http://dx.doi.org/10.1214/11-AIHP454}
{\path{doi:10.1214/11-AIHP454}}

\bibitem{add1}
Kallummil S, Kalyani S (2019) Noise statistics bblivious GARD for robust regression with sparse outliers. IEEE Trans Signal Process 67(2):383-398 
\newblock \href {http://dx.doi.org/10.1109/TSP.2018.2883025}
{\path{doi:10.1109/TSP.2018.2883025}}

\bibitem{Christmann2007Consistency}
Christmann A, Steinwart I (2007) Consistency and robustness of kernel based
regression. Bernoulli 13(3):799--819
\newblock \href {http://dx.doi.org/10.3150/07-BEJ5102}
{\path{doi:10.3150/07-BEJ5102}}

\bibitem{add3}
Huang D, Cabral R, Torre FDl (2016) Robust regression. IEEE Trans Pattern Anal Mach Intell 38(2):363-375 
\newblock \href {http://dx.doi.org/10.1109/TPAMI.2015.2448091}
{\path{doi:10.1109/TPAMI.2015.2448091}}

\bibitem{Zhang2018}
Zhang L, Zhou ZH (2018) $\ell_1$-regression with heavy-tailed distributions. in: Proceedings of Advances in Neural Information Processing Systems 

\bibitem{Yao1996Asymmetric}
Yao Q, Tong H (2007) Asymmetric least squares regression estimation: A
nonparametric approach. J Nonparametr Stat 6(4):273--292
\newblock \href {http://dx.doi.org/10.1080/10485259608832675]}
{\path{doi:10.1080/10485259608832675]}}

\bibitem{Chen2017Least}
Chen C, Li Y, Yan C, Guo J, Liu G (2017) Least absolute deviation-based robust
support vector regression. Knowledge-Based Syst 131(1):183--194
\href {http://dx.doi.org/10.1016/j.knosys.2017.06.009}
{\path{doi:10.1016/j.knosys.2017.06.009}}

\bibitem{Chen2015A}
Chen C, Yan C, Li Y (2015) A robust weighted least squares support vector
regression based on least trimmed squares. Neurocomputing 168(30):941--946
\newblock \href {http://dx.doi.org/10.1016/j.neucom.2015.05.031}
{\path{doi:10.1016/j.neucom.2015.05.031}}

\bibitem{Mangasarian2002Robust}
Mangasarian OL, Musicant DR (2002) Robust linear and support vector regression.
IEEE Trans Pattern Anal Mach Intell 22(9):950--955
\newblock \href {http://dx.doi.org/10.1109/34.877518}
{\path{doi:10.1109/34.877518}}

\bibitem{Huber1}
Huber PJ (2014) Robust statistics. Springer, New York
\newblock \href {http://dx.doi.org/10.1007/978-3-642-04898-2_594}
{\path{doi:10.1007/978-3-642-04898-2_594}}

\bibitem{Huber1964Robust}
Huber PJ (1964) Robust estimation of a location parameter. Ann Math Statist 35(1):73--101
\newblock \href {http://dx.doi.org/10.1214/aoms/1177703732}
{\path{doi:10.1214/aoms/1177703732}}

\bibitem{nipssvm}
Christmann A, Steinwart I (2007) How svms can estimate quantiles and the median.
In: Proceedings of Advances in Neural Information Processing Systems, pp 305--312

\bibitem{Omer2017Maximum}
Karal O (2017) Maximum likelihood optimal and robust support vector regression with
$lncosh$ loss function. Neural Netw 94(10):1--12
\newblock \href {http://dx.doi.org/10.1016/j.neunet.2017.06.008}
{\path{doi:10.1016/j.neunet.2017.06.008}}

\bibitem{Ren2018Correntropy}
Ren Z, Yang Y (2018) Correntropy-based robust extreme learning machine for
classification. Neurocomputing 313(11):74--84
\newblock \href {http://dx.doi.org/10.1016/j.neucom.2018.05.100}
{\path{doi:10.1016/j.neucom.2018.05.100}}

\bibitem{Kai2015Outlier}
Kai Z, Luo M (2015) Outlier-robust extreme learning machine for regression
problems. Neurocomputing 151(3):1519--1527
\newblock \href {http://dx.doi.org/10.1016/j.neucom.2014.09.022}
{\path{doi:10.1016/j.neucom.2014.09.022}}

\bibitem{Yang2018}
Yang L, Dong H (2018) Support vector machine with truncated pinball loss and its
application in pattern recognition. Chemometrics Intell Lab Syst 177(6):89--99
\newblock \href {http://dx.doi.org/10.1016/j.chemolab.2018.04.003}
{\path{doi:10.1016/j.chemolab.2018.04.003}}

\bibitem{Yang2020}
Yang L, Dong H (2019) Robust support vector machine with generalized quantile loss for classification and regression. Appl Soft Comput 81(8):105483
\newblock \href {http://dx.doi.org/10.1016/j.asoc.2019.105483}
{\path{doi:10.1016/j.asoc.2019.105483}}

\bibitem{Holland2017Robust}
Holland MJ, Ikeda K (2017) Robust regression using biased objectives. Mach Learn 106(4):1--37
\newblock \href {http://dx.doi.org/10.1007/s10994-017-5653-5}
{\path{doi:10.1007/s10994-017-5653-5}}

\bibitem{Lugosi2016Risk}
Lugosi G, Mendelson S (2020) Risk minimization by median-of-means tournaments. J Eur Math Soc 22(3):925-965
\newblock \href {http://dx.doi.org/10.4171/JEMS/937}
{\path{doi:10.4171/JEMS/937}}

\bibitem{catoni2009}
Catoni O (2009) High confidence estimates of the mean of heavy-tailed real random
variables. arXiv:0909.5366

\bibitem{Suykens2002Weighted}
Suykens JAK, Brabanter JD, Lukas L, Vandewalle J (2002) Weighted least
squares support vector machines: Robustness and sparse approximation. Neurocomputing 48(10):85--105
\newblock \href {http://dx.doi.org/10.1016/s0925-2312(01)00644-0}
{\path{doi:10.1016/s0925-2312(01)00644-0}}


\bibitem{Wang2014Robust}
Wang K, Zhong P (2014) Robust non-convex least squares loss function for regression
with outliers. Knowledge-Based Syst 71:290--302
\newblock \href {http://dx.doi.org/10.1016/j.knosys.2014.08.003}
{\path{doi:10.1016/j.knosys.2014.08.003}}

\bibitem{Zhao2010Robust}
Zhao Y, Sun J (2010) Robust truncated support vector regression. Expert Syst Appl 37(7):5126--5133
\newblock \href {http://dx.doi.org/10.1016/j.eswa.2009.12.082}
{\path{doi:10.1016/j.eswa.2009.12.082}}


\bibitem{chen2017neurocom}
Chen K, Lv Q, Lu Y, Dou Y (2016) Robust regularized extreme learning machine for regression using iteratively reweighted least squares. Neurocomputing 230(12):345--358
\newblock \href {http://dx.doi.org/10.1016/j.neucom.2016.12.029}
{\path{doi:10.1016/j.neucom.2016.12.029}}


\bibitem{Akoa2008Combining}
Dinh DP, Thi HAL, Akoa F (2008) Combining {DCA} ({DC} {A}lgorithms) and
interior point techniques for large-scale nonconvex quadratic programming.
Optim Methods Softw 23(4):609--629
\newblock \href {http://dx.doi.org/10.1080/10556780802263990}
{\path{doi:10.1080/10556780802263990}}

\bibitem{Yang2016A}
Yang L, Qian Y (2016) A sparse logistic regression framework by difference of
convex functions programming. Appl Intell 45(2):241--254
\newblock \href {http://dx.doi.org/10.1007/s10489-016-0758-2}
{\path{doi:10.1007/s10489-016-0758-2}}

\bibitem{YUILLE2002CCCP}
Yuille AL (2002) {CCCP} algorithms to minimize the {B}ethe and {K}ikuchi free
energies: {C}onvergent alternatives to belief propagation. Neural Comput 14(7):1691--1722
\newblock \href {http://dx.doi.org/10.1162/08997660260028674}
{\path{doi:10.1162/08997660260028674}}

\bibitem{Zhang2013robust}
Zhang Y, Sun Y, He R, Tan T (2013) Robust subspace clustering via half-quadratic
minimization. In: Proceedings of IEEE International Conference on Computer Vision, pp 3096--3103

\bibitem{Ran2014Half}
He R, Zheng W, Tan T, Sun Z (2014) Half-quadratic-based iterative minimization
for robust sparse representation. IEEE Trans Pattern Anal Mach Intell 36(2):261--275
\newblock \href {http://dx.doi.org/10.1109/TPAMI.2013.102}
{\path{doi:10.1109/TPAMI.2013.102}}

\bibitem{Feng2016Robust}
Feng Y, Yang Y, Huang X, Mehrkanoon S, Suykens JAK (2016) Robust support
vector machines for classification with nonconvex and smooth losses. Neural
Comput 28(6):1217--1247
\newblock \href {http://dx.doi.org/10.1162/NECO_a_00837}
{\path{doi:10.1162/NECO_a_00837}}

\bibitem{Li2017Sparse}
Li C, Zhou S (2017) Sparse algorithm for robust {LSSVM} in primal space.
Neurocomputing 275:2880--2891
\newblock \href {http://dx.doi.org/10.1016/j.neucom.2017.10.011}
{\path{doi:10.1016/j.neucom.2017.10.011}}

\bibitem{Xu2016Robust}
Xu G, Hu B, Principe JC (2016) Robust {C}-loss kernel classifiers. IEEE Trans Neural Netw Learn Syst 29(3):510--522
\newblock \href {http://dx.doi.org/10.1109/TNNLS.2016.2637351}
{\path{doi:10.1109/TNNLS.2016.2637351}}

\bibitem{Lai2013Improved}
Lai MJ, Xu Y, Yin W (2013) Improved iteratively reweighted least squares for
unconstrained smoothed $\ell$q minimization. SIAM J Numer Anal 51(2):927--957
\newblock \href {http://dx.doi.org/10.1137/110840364}
{\path{doi:10.1137/110840364}}

\bibitem{Green1984Iteratively}
Green PJ (1984) Iteratively reweighted least squares for maximum likelihood
estimation, and some robust and resistant alternatives. J R Stat Soc Ser B-Stat Methodol 46(2):149--192
\newblock \href {http://dx.doi.org/10.1111/j.2517-6161.1984.tb01288.x}
{\path{doi:10.1111/j.2517-6161.1984.tb01288.x}}

\bibitem{irlskbr}
Debruyne M, Christmann A, Hubert M, Suykens JAK (2010) Robustness of
reweighted least squares kernel based regression. J Multivar Anal 101:447--463
\newblock \href {http://dx.doi.org/10.1016/j.jmva.2009.09.007}
{\path{doi:10.1016/j.jmva.2009.09.007}}

\bibitem{add2}
Yi S, He Z, Cheung YM, Chen WS (2018) Unified sparse subspace learning via self-contained regression. IEEE Trans Circuits Syst Video Technol 28(10):2537-2550 
\newblock \href {http://dx.doi.org/10.1109/TCSVT.2017.2721541}
{\path{doi:10.1109/TCSVT.2017.2721541}}


\bibitem{4938676}
Deng W, Zheng Q, Chen L (2009) Regularized extreme learning machine. In: Proceedings of IEEE
Symposium on Computational Intelligence and Data Mining, pp 389--395
\newblock \href {http://dx.doi.org/10.1109/CIDM.2009.4938676}
{\path{doi:10.1109/CIDM.2009.4938676}}

\bibitem{Christmann2008Support}
Christmann A, Steinwart I (2008) Support vector machines. Springer, New York
\newblock \href {http://dx.doi.org/10.1007/978-0-387-77242-4}
{\path{doi:10.1007/978-0-387-77242-4}}

\bibitem{Ref1006}
Hampel FR, Ronchetti EM, Rousseeuw PJ, Stahel WA (1986) Robust
statistics: The approach based on influence functions. Wiley, New York
\newblock \href {http://dx.doi.org/10.2307/1269782}
{\path{doi:10.2307/1269782}}

\bibitem{pinsvm}
Huang X, Shi L, Suykens JAK (2014) Support vector machine classifier with pinball loss. IEEE Trans Pattern Anal Mach Intell 36(5):984--997
\newblock \href {http://dx.doi.org/10.1109/TPAMI.2013.1786}
{\path{doi:10.1109/TPAMI.2013.178}}

\bibitem{Blake1998UCI}
Dua D, Graff C (2019) {UCI} machine learning repository.
(\url{http://archive.ics.uci.edu/ml})

\bibitem{Ref1010}
Deng N, Tian Y, Zhang C (2012) Support vector machines: Optimization based theory, algorithms and extensions. CRC Press, Boca Raton

\bibitem{Crambes2013}
Crambes C, Gannoun A, Henchiri Y (2013) Support vector machine quantile regression approach for functional data: {S}imulation and application studies. J Multivar Anal 121(11):50--68
\newblock \href {http://dx.doi.org/10.1016/j.jmva.2013.06.004}
{\path{doi:10.1016/j.jmva.2013.06.004}}
\end{thebibliography}
\end{document}